\documentclass{article} % For LaTeX2e
\usepackage[table]{xcolor}        % colors
\usepackage{iclr2025_conference,times}
\iclrfinalcopy 

\usepackage{amsmath,amsfonts,bm}

\def\eqref#1{equation~\ref{#1}}
\def\1{\bm{1}}

\DeclareMathAlphabet{\mathsfit}{\encodingdefault}{\sfdefault}{m}{sl}
\SetMathAlphabet{\mathsfit}{bold}{\encodingdefault}{\sfdefault}{bx}{n}

\newcommand{\R}{\mathbb{R}}

\usepackage{url}
\usepackage{arydshln}
\usepackage[utf8]{inputenc} % allow utf-8 input
\usepackage[T1]{fontenc}    % use 8-bit T1 fonts
\usepackage{graphicx}
\usepackage{booktabs}       % professional-quality tables
\usepackage{amsfonts}       % blackboard math symbols
\usepackage{nicefrac}       % compact symbols for 1/2, etc.
\usepackage{microtype}      % microtypography
\usepackage[numbers,sort&compress]{natbib}
\usepackage{amssymb}
\usepackage{algpseudocode}
\usepackage{algorithm}
\usepackage{multirow}
\usepackage{arydshln}
\usepackage{pifont}
\usepackage{array}
\usepackage{wrapfig}
\usepackage{subcaption}
\usepackage[font=small]{caption}
\usepackage{wrapfig}
\usepackage{enumitem}
\usepackage{amsmath}
\usepackage{mathrsfs} % for script math symbols
\usepackage{tcolorbox}
\newcommand{\VarSty}[1]{\textnormal{\ttfamily\color{blue!90!black}#1}\unskip}
\newcommand{\var}{\texttt}
\newcommand{\StreamChat}{\texttt{\normalsize S\footnotesize TREAM\normalsize C\footnotesize HAT}}
\newcommand{\StreamBench}{\texttt{\normalsize S\footnotesize TREAM\normalsize B\footnotesize ENCH}}
\newcommand{\StreamChats}{\texttt{\scriptsize S\tiny TREAM\scriptsize C\tiny HAT}}
\newcommand{\StreamBenchs}{\texttt{\scriptsize S\tiny TREAM\scriptsize B\tiny ENCH}}
\newcommand{\cmark}{\textcolor{green!40!black}{\ding{51}}}
\newcommand{\xmark}{\textcolor{red}{\ding{55}}}

\usepackage{hyperref}
\definecolor{c1}{HTML}{177cb0}
\definecolor{c2}{HTML}{065279}
\hypersetup{colorlinks=true,
    linkcolor=c1,        % color of internal links
    citecolor=c2,        % color of citations 
    filecolor=magenta}

\usepackage[capitalize]{cleveref}

\title{Streaming Video Understanding and Multi-round Interaction with Memory-enhanced Knowledge} 

\author{
\centerline{Haomiao Xiong$^1$\thanks{Equal contribution.},~~Zongxin Yang$^2$\footnotemark[1],~~Jiazuo Yu$^1$,~~Yunzhi Zhuge$^1$\thanks{Corresponding author (\texttt{zgyz@dlut.edu.cn}).}}\\ 
\centerline{\textbf{Lu Zhang}$^1$,~~\textbf{Jiawen Zhu}$^1$,~~\textbf{Huchuan Lu}$^1$}\\ 
\centerline{$^1$Dalian University of Technology, $^2$Harvard University}
}

\begin{document}

\maketitle

\begin{abstract}

Recent advances in Large Language Models (LLMs) have enabled the development of Video-LLMs, advancing multimodal learning by bridging video data with language tasks. However, current video understanding models struggle with processing long video sequences, supporting multi-turn dialogues, and adapting to real-world dynamic scenarios. To address these issues, we propose $\StreamChat$, a training-free framework for streaming video reasoning and conversational interaction. $\StreamChat$ leverages a novel hierarchical memory system to efficiently process and compress video features over extended sequences, enabling real-time, multi-turn dialogue. Our framework incorporates a parallel system scheduling strategy that enhances processing speed and reduces latency, ensuring robust performance in real-world applications. Furthermore, we introduce $\StreamBench$, a versatile benchmark that evaluates streaming video understanding across diverse media types and interactive scenarios, including multi-turn interactions and complex reasoning tasks.  Extensive evaluations on $\StreamBench$ and other public benchmarks demonstrate that $\StreamChat$ significantly outperforms existing state-of-the-art models in terms of accuracy and response times, confirming its effectiveness for streaming video understanding. Code is available at \href{https://github.com/hmxiong/StreamChat}{StreamChat}.

\end{abstract}

\section{Introduction}\label{introduction}
\begin{wrapfigure}{rd}{0.42\textwidth}
  \vspace{-5mm}
  \centering
  \setlength{\abovecaptionskip}{0cm}
  \includegraphics[width=0.42\textwidth]{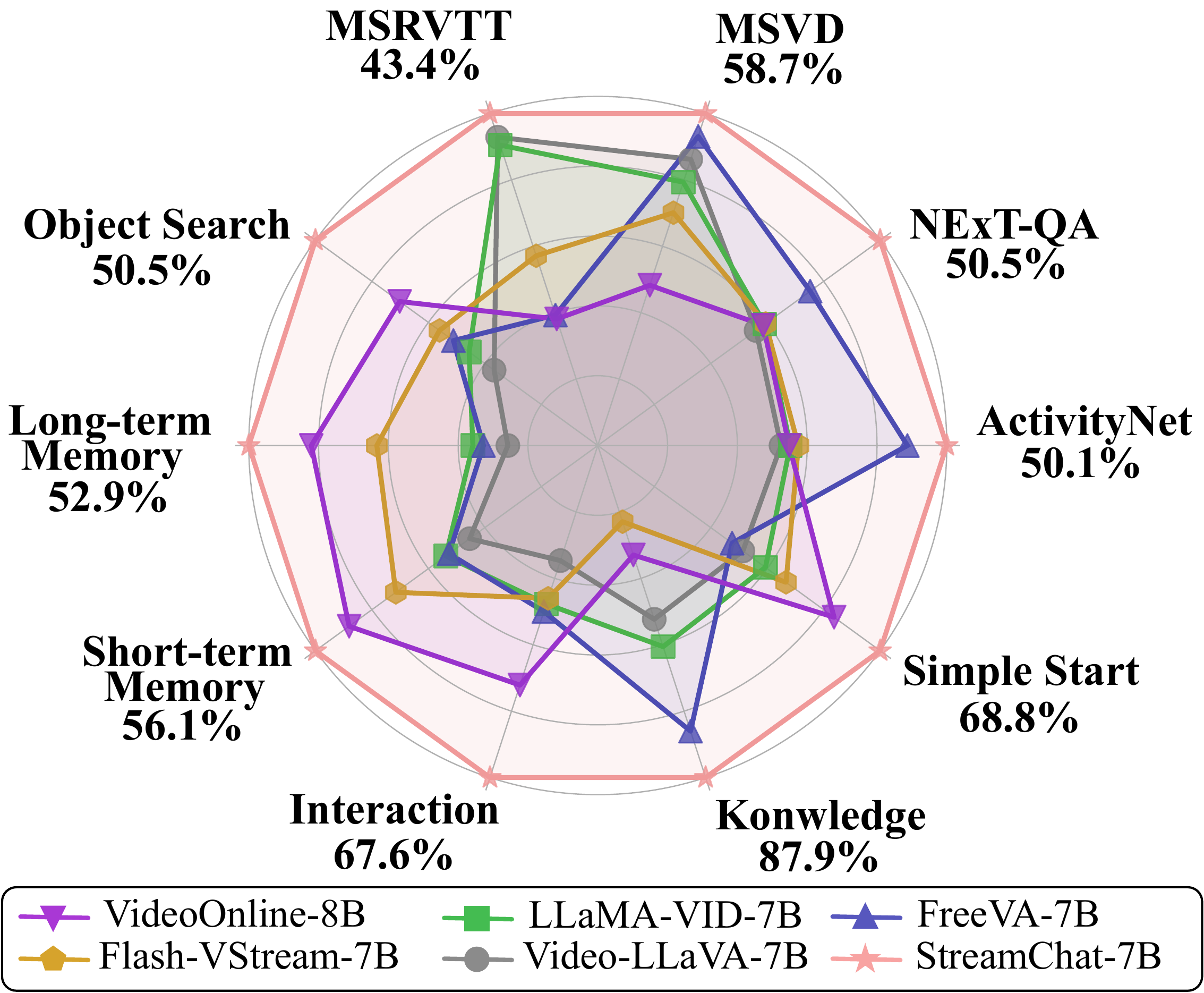}
  \vspace{-3mm}
  \caption{
  \textbf{Performance comparison} between $\StreamChat$ and previous Video-LLMs.}
  \label{figure:performance}
  \vspace{-7mm}
\end{wrapfigure}

Recent advancements in Large Language Models (LLMs) \cite{openai2022chatgpt,li2023llama,dubey2024llama} have led to the development of Video-LLMs~\cite{lin2023video,jin2024chat,maaz2023video,song2024moviechat,wu2024freeva,xu2024pllava}, which aim to interpret visual scenes, actions, and narratives. These models represent significant progress in multimodal learning by bridging video data and language-based tasks, with applications spanning from content analysis to human-robot interaction~\cite{xu2024pllava}.

% Despite these advancements, current offline methods primarily treat videos as static clips and rely on single-round dialogues by incorporating vision information through projection layers~\cite{lin2023video, wu2024freeva} or cross-attention structures~\cite{li2023llama}. These methods struggle with computational challenges when processing extended video sequences, often failing to compress lengthy video features within constrained memory capacities~\cite{zhang2024flash}. Furthermore, their inability to support multi-turn dialogues limits adaptability to interactive user needs and can result in the omission of critical information due to video sampling methods (\textit{cf.} Fig.~\ref{figure:teaser}(a)).

\begin{figure*}[t]
  \centering
  \vspace{-3mm}
  \setlength{\abovecaptionskip}{0cm}
  \includegraphics[width=1.0\textwidth]{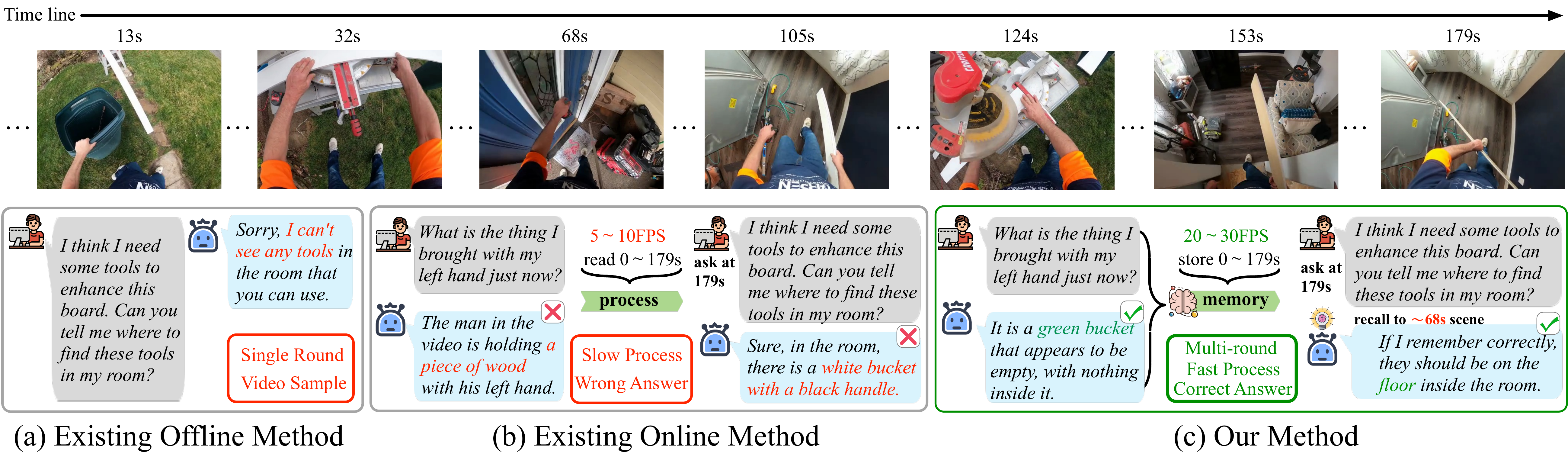}
  \vspace{-3mm}
  \caption{\textbf{The comparisons between StreamChat and other methods} (\S\ref{introduction}). Offline methods process entire videos, leading to information loss and limited to a single interaction. Previous online methods~\cite{chen2024videollm, zhang2024flash} enable multi-round interactions but still suffer from slow processing and answer correctly. The proposed method achieves real-time video processing, improving the efficiency and accuracy with memory support.}
  \label{figure:teaser}
  \vspace{-5mm}
\end{figure*}

Despite these advancements, current offline models primarily process videos as static clips and rely on single-turn dialogues, incorporating visual information through mechanisms like projection layers~\cite{lin2023video, wu2024freeva} or cross-attention structures~\cite{li2023llama}. However, these models encounter computational bottlenecks when handling extended video sequences, often struggling to compress lengthy video features within limited memory resources~\cite{zhang2024flash}. Additionally, their inability to support multi-turn dialogues reduces adaptability for interactive scenarios, and key information may be lost due to insufficient video sampling methods (\textit{cf.} Fig.~\ref{figure:teaser}(a)).

% In response to these limitations, online models~\cite{chen2024videollm, zhang2024flash} have been developed to handle long video sequences using memory-based methods and temporally aligned instruction-tuning. While they allow multi-round user interactions, as illustrated in Fig.~\ref{figure:teaser}(b), these models still face challenges in rapid processing speeds and maintaining stable performance across novel scenarios, which are crucial in real-time applications like robotic navigation and human-robot interaction.

To address these issues, online
 models~\cite{chen2024videollm, zhang2024flash} have emerged. They utilize memory-based approaches and temporally aligned instruction-tuning to process long videos and enable multi-round interactions (\textit{cf.} Fig.~\ref{figure:teaser}(b)). While these models allow dynamic user interactions, they still face challenges maintaining rapid processing speeds and consistent performance across unfamiliar scenarios—critical factors in real-time applications like robotic navigation and human-robot collaboration.

% Therefore, online models~\cite{chen2024videollm, zhang2024flash} are proposed to solve long video understanding by applying a memory-based method or temporally aligned instruction-tuning and to handle multi-round user requests.
% However, as shown in Fig.~\ref{figure:teaser}(b), existing online methods still \textit{struggle with fast video processing and stable performance in new scenarios}.
% Besides, the \textit{huge training resource overhead} remains a challenge for researchers with limited computation and data resources.
% Therefore, memory-based methods~\cite{song2024moviechat, zhang2024flash} and agents methods~\cite{wang2024videotree} first try to solve the source budget caused by long video frames but still fail to achieve real-time video processing.
% Lastly, the user’s questions may continuously change as the video progresses, but existing methods lack the multi-turn dialogue capability to adapt to the user’s dynamic information needs.
% The emergence of models has driven the development of test data, and a lot of benchmarks~\cite{yu2019activitynet,xiao2021next,xu2017video,xu2016msr}  have emerged to verify the video understanding capabilities of models.

To overcome these limitations, we propose $\StreamChat$, a training-free framework for streaming video understanding that offers three key innovations: \textbf{(i) Training-free adaptability}, allowing it to efficiently process videos of various types and lengths without resource-heavy training. This makes $\StreamChat$ suitable for both online and offline video processing while maintaining stable performance across diverse scenarios. \textbf{(ii) Hierarchical memory storage}, which manages and compresses video information over long sequences. It integrates short-term memory for tracking ongoing events, long-term memory for retaining past events in compressed form, and dialogue memory to maintain conversational history, ensuring continuous and coherent dialogue understanding. \textbf{(iii) Optimized system scheduling}, which improves model inference efficiency by processing tasks in parallel across three threads: the \textit{selective frame stacking} thread identifies and removes redundant frames, the \textit{memory formation} thread updates and refines memory information, and the \textit{contextual summarization} thread handles user requests and generates responses in real-time.

We evaluate $\StreamChat$ on existing benchmarks~\cite{xu2016msr, mangalam2024egoschema, yu2019activitynet, xiao2021next, xu2017video} and identify their two major shortcomings: \textbf{(\romannumeral1) Short and monotonous video content}, which fails to capture the complexity of real-world streaming media, and \textbf{(\romannumeral2) Simplistic, single-round questions}, which do not test the model's ability to engage in multi-turn dialogue or complex reasoning.

To address these deficiencies, we introduce $\StreamBench$, a comprehensive benchmark designed for streaming video understanding. It includes a diverse array of video content such as egocentric videos, web videos, and movie scenes, paired with text annotations that simulate multi-round interactions. \textit{In terms of video selection}, we perform rigorous manual curation from large datasets to ensure both high-quality content and a broad range of categories. \textit{In terms of questions}, we design six distinct types of queries, probing various dimensions of the model’s reasoning abilities, from simple factual retrieval to complex inference. Compared to previous benchmarks, $\StreamBench$ not only evaluates the accuracy of model responses but also incorporates latency metrics, which are essential for assessing performance in real-time applications. This comprehensive evaluation framework offers a more realistic and reliable measure of model robustness and practical utility.

% Our main contributions can be summarized as the following parts:
In summary, our key contributions are as follows:
\begin{itemize}[leftmargin=*,itemsep=0pt]
% \item We launch $\StreamChat$, a training-free agent that can efficiently perform streaming video understanding. The hierarchical memory storage and the system scheduling strategy are innovatively designed to supply the agent with abundant memories and real-time video processing capability.
% \item We introduce $\StreamBench$, a novel benchmark to evaluate streaming video understanding. It incorporates multi-round dialogue and diverse question formats, simulating real-world interactions between Video-LLMs and users, enabling a comprehensive assessment of model performance.
% \item Extensive experiments are conducted to demonstrate that our StreamChat achieves state-of-the-art performance in both offline and online benchmarks.
% The results indicate that our StreamChat achieves 64.7\% accuracy on StreamBench, surpassing the previous best online method by 8.3\%, while further improving it by 2.5\% on average on offline benchmarks.
% \item Extensive experiments are conducted to demonstrate that our memory approach can assist the model achieve state-of-the-art performance in both offline and online benchmarks.

\item We propose $\StreamChat$, a training-free method for streaming video understanding. Its novel $\textit{hierarchical memory storage}$ and $\textit{system scheduling}$ strategy enables robust memory management, real-time video processing, and multi-round interaction capabilities. These features ensure precise and efficient response generation, catering to the dynamic nature of video contexts.

\item We introduce $\StreamBench$, the first comprehensive benchmark to evaluate streaming video understanding models. This benchmark simulates real-world interactions through multi-turn dialogues and diverse question formats, offering a detailed assessment of model performance.

% \item We introduce $\StreamBench$, the first comprehensive benchmark designed to evaluate streaming video understanding. It simulates real-world interactions through multi-turn dialogue and diverse question formats, offering a thorough assessment of model performance.

\item $\StreamChat$ sets new benchmarks (\textit{cf.} Fig.~\ref{figure:performance}), delivering a 64.7\% accuracy on $\StreamBench$ for online settings, which is an 8.3\% improvement over the previous best. In offline scenarios, it outperforms the state-of-the-art method by an average of 2.5\% across four public benchmarks. 
% Extensive experiments show that $\StreamChat$ sets new state-of-the-art benchmarks in both offline and online settings (\textit{cf} Fig.~\ref{figure:teaser}). It achieves a 64.7\% accuracy, surpassing the previous best online method on by 8.3\% $\StreamBench$ while improving offline performance by an average of 2.5\% on four benchmarks.
 
% \item We evaluate the performance of the $\StreamChat$ in terms of efficiency. During the video processing, our method is able to achieve a real-time speed of 30 FPS, which is 6 times faster than previous methods, and maintain a text generation latency of less than 0.9s.
\item In terms of efficiency, $\StreamChat$ achieves a processing speed of 32 FPS, marking a sixfold increase over existing methods. Additionally, it maintains text generation latency under 0.9 seconds, showcasing significant advances in interactive video processing.
% $\StreamChat$ reaches processing speeds of 30 FPS, six times faster than existing methods, and maintains a text generation latency of less than 0.9 seconds, demonstrating significant advances in real-time video processing capabilities.
\end{itemize}

\section{Collection and Composition of StreamBench}\label{benchmark}
\subsection{Video Collection}
\begin{figure*}[t]
  \centering
  \setlength{\abovecaptionskip}{0cm}
  \includegraphics[width=0.975\textwidth]{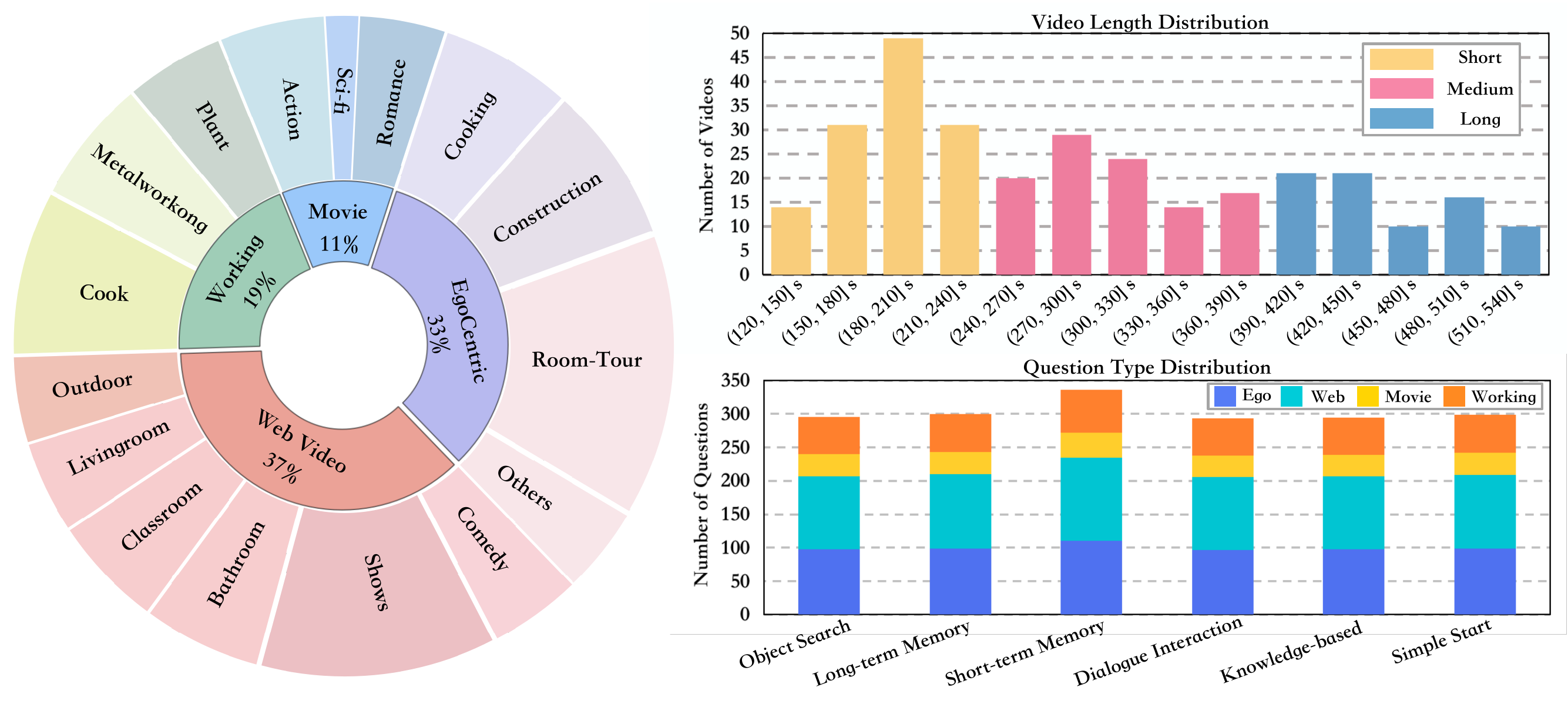}
  % \vspace{-4mm}
  % \vspace{-4mm}
  \caption{
  \textbf{Benchmark overview} (\S\ref{benchmark}). Our benchmark covers 4 key domains and 16 sub-class video types. These videos exhibit a broader distribution of length, with 6 different types that are evenly distributed.}
  %The number of 6 types of test questions is evenly distributed across different video types.
  \vspace{-4mm}
  \label{figure:StreamingBench}
\end{figure*}

\begin{wraptable}{r}{0.43\textwidth}
\vspace{-13pt}
\centering
\scriptsize
\renewcommand\arraystretch{1.0}
\setlength{\tabcolsep}{3pt} % 压缩列间距
\captionof{table}{\textbf{Comparisons of different benchmarks}. MRI denotes multi-round interactions.}
% The videos used in StreamBench are a mixture of Web, Ego, Movie, and Working. What's more, the average length of each video is longer than others. }
\vspace{-8pt}
\begin{tabular}{|l||ccccc|}
    \hline
     \rowcolor{gray!45} Benchmark  &  \textbf{MR} &  \textbf{Avg}  &  \textbf{Total}   &  \textbf{Video}  &  \textbf{QA}   \\
    \hline\hline
     MSVD~\cite{xu2017video}              & \xmark &  10s              &  1.4h             &  Web              &  Desc.        \\
     MSRVTT~\cite{xu2016msr}              & \xmark &  15s              &  12.5h               &  Web              &  Desc.        \\
     ActivityNet~\cite{yu2019activitynet} & \xmark &  112s          &  25h                 &  Web               &  Desc.        \\
     Next-QA~\cite{xiao2021next}          & \xmark &  40s           &  11h                 &  Web               &  Temporal     \\
     MovieChat~\cite{song2024moviechat}   & \xmark &  213s               &  9h                  &  Movie             &  Movie        \\
    \hline
    $\StreamBenchs$    & \cmark &  \textbf{270s} &  \textbf{25h} &  \textbf{Mix} &  \textbf{Online} \\
    \hline
    % \multirow{2}{*}{\scriptsize \textbf{StreamBench}}    & \multirow{2}{*}{\scriptsize \textbf{270s}} & \multirow{2}{*}{\scriptsize 24h} & \multirow{2}{*}{\scriptsize Mix} & \multirow{2}{*}{\scriptsize Online} \\
    % ~                                       & ~                               & ~                    & ~                     & ~                                                                             \\
    % \bottomrule
\end{tabular}
\label{tab:benchmark_comparison}
\vspace{-15pt}
\end{wraptable}
Previous video understanding benchmarks~\cite{xu2017video, xu2016msr, yu2019activitynet, mangalam2024egoschema, fu2024video, zhou2024mlvu}  primarily focus on offline scenarios, where all video frames and user questions are provided to the model simultaneously for generating answers. 
% There is an urgent need to find more suitable methods to assess a model’s ability to understand online scenarios. 
% Thus, we introduce $\StreamBench$, a benchmark mainly designed to simulate online video scenarios.
To find a more suitable method to assess a model's ability to understand online scenarios, we introduce $\StreamBench$, a benchmark mainly designed to simulate online video scenarios.
There are two distinct differences compared with other video understanding benchmarks.
\textbf{(i)} \textbf{Diverse video curation}: we collect four major domains and sixteen sub-classes of video sources, including egocentric videos, web videos, working videos, and movies as the database of the benchmark.
Each type has its unique characteristics and challenges, which can verify the stability and reliability of the model in a wide range of application scenarios.
\textbf{(ii)} \textbf{Crafted query types}: we design six types of questions to meet the specific needs of online video understanding and ensure that these types of questions appear once in a single video, forming a multi-round dialogue.
This section introduces how we collect videos and construct annotations. 
More details about the diversity and distribution of our benchmark are shown in Fig.~\ref{figure:StreamingBench}.

% \vspace{-9pt}
\textbf{Data Sources}. 
\label{sec2.2}
% Various scenarios are considered when collecting videos.
% During data selection, we mainly consider the 
In selecting videos for our benchmarks, we prioritized diversity in type and length to maintain high data quality. Our primary sources are the EgoSchema~\cite{mangalam2024egoschema} and YouTube-8M~\cite{abu2016youtube} datasets. EgoSchema offers a rich array of both indoor and outdoor scenes, providing an extensive range of egocentric perspectives and actions, which aligns perfectly with our experimental needs. From YouTube-8M, which features a comprehensive internet-sourced collection spanning over 4,000 classes, we filtered to procure high-quality web, work-related, and cinematic videos. This diverse selection framework ensures our model is tested against a broad spectrum of real-world scenarios.
% Through research, we select EgoSchema~\cite{mangalam2024egoschema} and YouTube-8M~\cite{abu2016youtube} as the original video sources.
% Among them, EgoSchema~\cite{mangalam2024egoschema} provides a diverse range of indoor and outdoor scenes, making it an ideal fit for our requirements. This dataset allows us to capture a wide variety of real-world perspectives and actions from an egocentric view.
% The YouTube-8M~\cite{abu2016youtube} dataset encompasses a vast collection of videos sourced from the Internet, covering over 4,000 distinct classes of content, offering a wide range of visual data. Thus, we obtain high-quality 
% web videos, working videos, and movie data from it by further filtering.

\textbf{Filtering videos}. It is a crucial step to ensure the quality of the videos used in the benchmark. 
To achieve this, our data filtering pipeline consists of machine and human selection. 
Firstly, a multi-modal language model~\cite{zhang2024long} is utilized to classify the original data. The categories of videos are provided by data source, we feed them with the videos to the machine and make it select the category of the video. 
Secondly, human judgment is required to assess the redundancy: the change of scenes in videos.
Some static video content (e.g., ego view of drawing, watching TV) and high-noise data from web videos (e.g., video games, advertisements) are removed according to human judgment.
Finally, StreamBench consists of \textbf{306 videos} with a total duration of \textbf{24.8 hours} and an average of \textbf{4.5 minutes} each, offering a comprehensive collection of videos from different categories and lengths.

\subsection{Construction of Tasks and Annotations}
 We have crafted six distinct tasks with annotations to simulate the conversation between the agent and the human. Each task corresponds to a different real-world scenario, ensuring comprehensive coverage of potential communication contexts.
\begin{itemize}[leftmargin=*,itemsep=0pt]
\item \textbf{Object Search (OS):} 
Challenges include accurately describing an object's position in a video. The task conditions are that the object must appear for less than 5 seconds and the interval from its appearance to the user's request should exceed 30 seconds, enhancing the difficulty of the search.
% Accurately describing the position of a specific object within a lengthy timeline is highly challenging for video understanding multi-modal large language models. 
% For this benchmark, we set the time interval between the target's appearance and the user’s request must be > 30 s, and the object’s appearance time must be < 5 s to ensure the search difficulty.
\item \textbf{Long-term Memory Search (LM):} 
This task assesses the model's memory by requiring recall of events appearing for more than 5 seconds, with a delay exceeding 1 minute from the event's end to the user's query, testing long-term memory retention.
% The primary objective of this type of query is to evaluate the model's memory and its ability to recall events stored in memory.
% In this case, we set the target event to appear > 5s and the time interval between the end of the target’s appearance and the user’s request must be > 1 minute.
% The model must recall memory accurately to answer the question.
\item \textbf{Short-term Memory Search (SM):}  To simulate the user's interest in recent events, this task sets the interval from event completion to the user’s query at less than 20 seconds, evaluating the model's response to recent activities.
% This type of question is intended to simulate the user's interest in recent events.
% We set the time interval between the end of the event and the user’s request to be < 20 seconds. This setup tests the model’s sensitivity to recent events and associative capabilities.
\item \textbf{Conversational Interaction (CI):}
Sometimes the answer to a user’s current question is closely related to conversation history. 
Therefore, the model must memorize conversation records and retrieve the most relevant text from the memory as contextual support. 
This type is designed to simulate multi-turn dialogue scenarios. 
We set the dialogue information associated with the user’s current request to come from any previous conversation, with an interval of more than 2 dialogues.
\item \textbf{Knowledge-based Question Answering (KG):}
This type of question evaluates the model’s internal knowledge, which is retained by the base large language models. 
In this benchmark, we set the questions must be related to the events or objects occurring in the video so that it can simulate scenarios where users have a specific need to understand background or encyclopedic knowledge.
\item \textbf{Simple Factual (SF):}
This type of question focuses on friendly dialogue starting between the user and the model. 
Therefore, they must be asked within 30s after the beginning of the video.
Although the question is simple, the model needs to remember things in the short term to answer correctly.
\end{itemize}
To ensure the quality of the annotations, we additionally assign different workers to perform human feedback for manual annotation. 
%This step of manual annotation, along with automated data collection, forms our semi-automated benchmark construction pipeline (details in Appen.$\S$~\ref{sec_app:collection_ipeline}).
%Apart from manual annotation, we set reviewers to check the annotations and provide feedback to ensure the quality of the annotations.
%The entire semi-automatic benchmark construction process, including video collection and text annotation is detailed in Appen.$\S$~\ref{sec_app:collection_ipeline}. 
The human feedback step needs to focus on three parts: (1) check the question formats are correct and diverse, (2) ensure the expressions are clear and consistent with the video, and (3) remove sensitive topics such as those questions related to nationality or politics. 
These steps of manual annotation and feedback, along with multi-modal large language model assisted video collection, form our semi-automated benchmark construction pipeline (Appen.~$\S$\ref{sec_app:collection_ipeline}).
Finally, $\StreamBench$ contains \textbf{1.8K} high-quality QA pairs.
The distribution of these annotations is shown in Fig.~\ref{figure:StreamingBench}.
Some examples from the benchmark that offer an intuitive observation of our annotation results and formats are shown Appen.~$\S$\ref{sec_app:visiualize}.
% Appen.$\S$~\ref{sec_app:visiualize} shows some examples from the benchmark to offer an intuitive observation for our annotation results and formats.

\section{StreamChat}\label{streamchat}
% (整体围绕streaming的特点) pipeline -> 分点问题：buffer不能无限增长 -> 开始介绍memory -> 考虑到系统的特性
% Based on the idea of the Atkinson-Shiffrin model~\cite{atkinson1968proposed} for memory formation, storage, and retrieval, the hierarchical memory system addresses the storage and access of streaming video information by constructing a tree structure and search algorithms.
% \begin{table}[tbh]
%     \centering
%     \begin{tabular}{lccccc}
%         \toprule
%         \textbf{Method} & \textbf{Streaming} & \textbf{Training-free} & \textbf{Memory} & \textbf{Real-time} & \textbf{Low Latency} \\
%         \midrule
%         Video-LLaVA          & \xmark  & \xmark  & \xmark  & \xmark  & \xmark  \\
%         MovieChat            & \xmark  & \cmark  & \cmark  & \xmark  & \xmark  \\
%         Video-online         & \cmark  & \xmark  & \xmark  & \xmark	 & \xmark  \\
%         Flash-VStream        & \cmark  & \xmark  & \cmark  & \xmark	 & \xmark  \\
%         $\StreamChat$        & \cmark  & \cmark  & \cmark  & \cmark	 & \cmark  \\
%         \bottomrule
%     \end{tabular}
%     \caption{Comparisons of recent video understanding Multi-modal Large Language Models. Our \textit{training-free} method achieves processing video in \textit{real-time} and generates a response with \textit{low latency}.}
%     \label{tab:model_comparison}
% \end{table}
\setlength{\intextsep}{1pt}
\setlength{\columnsep}{7pt} 
\begin{wraptable}{r}{0.35\textwidth}
\renewcommand\arraystretch{1.0}
  \vspace{-4pt}
 
    \centering
    \renewcommand{\arraystretch}{0.8} % 压缩行间距
    \caption{\textbf{Comparisons of recent video-mllms}. Our \textit{streaming} (S.) method with \textit{memory} (M.) achieves processing video in \textit{real-time} (R.) and generates a response with \textit{low latency} (L.).}
    \vspace{-2mm}
    \begin{tabular}{|l||
    >{\centering\arraybackslash}p{0.13cm}
    >{\centering\arraybackslash}p{0.13cm}
    >{\centering\arraybackslash}p{0.13cm}
    >{\centering\arraybackslash}p{0.13cm}|}
    
        \hline
        \rowcolor{gray!45} \scriptsize Method & \scriptsize \textbf{S.} & \scriptsize \textbf{M.} & \scriptsize \textbf{R.} & \scriptsize \textbf{L.} \\
        \hline \hline
        \scriptsize Video-LLaVA~\cite{lin2023video}          & \xmark  & \xmark  & \xmark  & \xmark  \\
        \scriptsize MovieChat~\cite{song2024moviechat}              & \xmark  & \cmark  & \xmark  & \xmark  \\
        \scriptsize Video-online~\cite{chen2024videollm}         & \cmark  & \xmark  & \xmark  & \xmark  \\
        \scriptsize Flash-VStream~\cite{zhang2024flash}        & \cmark  & \cmark  & \xmark  & \xmark  \\
        \scriptsize {\StreamChats}        & \cmark  & \cmark  & \cmark  & \cmark  \\
        \hline
    \end{tabular}
    % \vspace{-5mm}
    \label{tab:charastic_comparison}
\end{wraptable}
% \begin{wraptable}{r}{0.32\textwidth}
%   \vspace{-4pt}
%     \centering
%     \renewcommand{\arraystretch}{0.4} % 压缩行间距
%     % \setlength{\tabcolsep}{1pt} % 压缩列间距
%     \caption{Comparisons of recent video MLLMs. Our \textit{streaming} (S.) method with \textit{memory} (M.) achieves processing video in \textit{real-time} (R.) and generates a response with \textit{low latency} (L.).}
%     \vspace{-6pt}
%     \begin{tabular}{l
%     >{\centering\arraybackslash}p{0.13cm}
%     >{\centering\arraybackslash}p{0.13cm}
%     >{\centering\arraybackslash}p{0.13cm}
%     >{\centering\arraybackslash}p{0.13cm}}
    
%         \toprule
%         \scriptsize Methods  & \scriptsize \textbf{S.} & \scriptsize \textbf{M.} & \scriptsize \textbf{R.} & \scriptsize \textbf{L.} \\
%         \midrule
%         \scriptsize Video-LLaVA~\cite{lin2023video}          & \xmark  & \xmark  & \xmark  & \xmark  \\
%         \scriptsize MovieChat~\cite{song2024moviechat}              & \xmark  & \cmark  & \xmark  & \xmark  \\
%         \scriptsize Video-online~\cite{chen2024videollm}         & \cmark  & \xmark  & \xmark  & \xmark  \\
%         \scriptsize Flash-VStream~\cite{zhang2024flash}        & \cmark  & \cmark  & \xmark  & \xmark  \\
%         \scriptsize {\StreamChats}        & \cmark  & \cmark  & \cmark  & \cmark  \\
%         \bottomrule
%     \end{tabular}
%     % \vspace{-5mm}
%     \label{tab:charastic_comparison}
% \end{wraptable}
% \textbf{Overview}. 
Given streaming broadcast video and timestamped questions as input, $\StreamChat$ is designed to efficiently perform reasoning and deliver accurate answers across multiple rounds. 
Building upon LongVA~\cite{zhang2024long} (Appen.~$\S$\ref{sec_app:selection}) as a foundational Video-LLM, our design incorporates two key components: 
% \textit{hierarchical memory storage} ($\S$\ref{sec:3.1}) and \textit{system scheduling} ($\S$\ref{sec:3.2}). The former leverages long-term, short-term, and dialogue memories to compress and store long video sequences within limited resources, preventing unbounded buffer growth. And the latter decouples video feature extraction from memory updates, enabling efficient reasoning.
% \textcolor{gray}{
a \textit{hierarchical memory storage} system ($\S$\ref{sec:3.1}) that leverages long-term, short-term, and dialogue memories to compress and manage extensive video sequences within constrained resources, thereby facilitating effective video-content reasoning; and a \textit{system scheduling}  strategy ($\S$\ref{sec:3.2}) that decouples video feature extraction from memory updates, thereby preventing unbounded buffer growth as the input video frames increase.
% } 
% These innovations collectively enhance video processing speed and significantly reduce the latency between user queries and the initiation of text generation.
 % \textcolor{blue}{more details in Appen.~$\S$\ref{sec_app:selection}.}
% StreamChat is a streaming video reasoning agent designed to efficiently compress and store long video sequences within limited resource. When provided with a video and a question, it extracts the most relevant information from memory using advanced retrieval algorithms. This retrieved data is then used as context to support accurate reasoning and answering.
% Furthermore, we decouple video analysis from memory updates, utilizing thread scheduling to improve inference efficiency.
% As shown in Fig.~\ref{figure:overview}, we address the challenge of unbounded buffer growth, which can slow down retrieval times as relevant visual data accumulates. To mitigate this, we introduce a hierarchical storage structure optimized for ultra-long buffers, significantly reducing VRAM usage while maintaining fast data access.
% As a result, this approach not only accelerates the speed of video processing but also achieves lower latency between user requests and the beginning of text generation.

\begin{figure*}[t]
  \centering
  \setlength{\abovecaptionskip}{0cm}
  \vspace{-4mm}
  \includegraphics[width=0.975\textwidth]{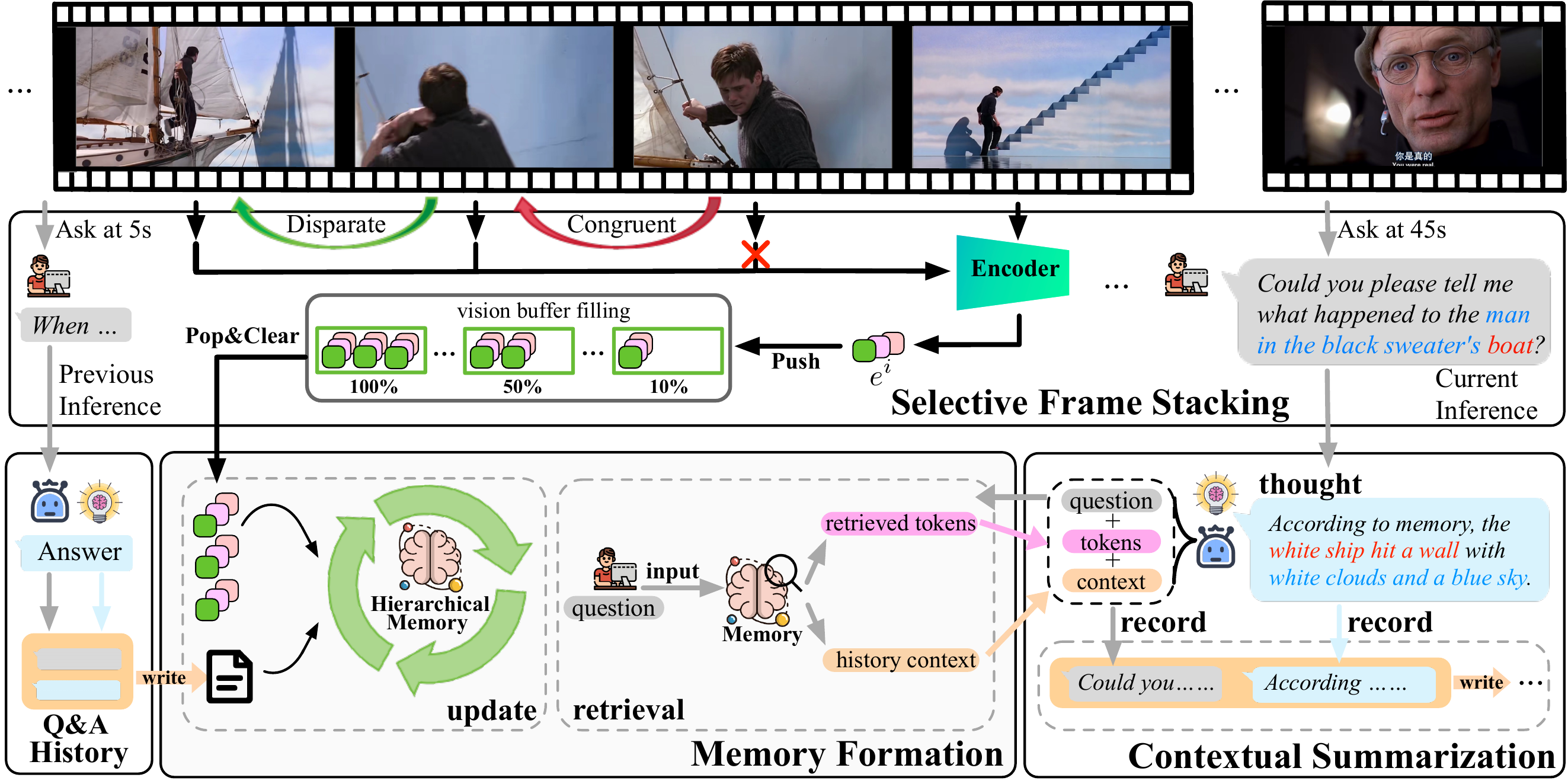}
  % \vspace{1mm}
  \caption{
  \textbf{Overview of StreamChat} (\S\ref{streamchat}), which
   comprises three main components: \textbf{(i) Selective frame stacking}, which prepares vision features for processing, including encoding frames and filling the vision buffer; \textbf{(ii) Memory formation}, where vision features are organized into structured memory; \textbf{(iii) Contextual summarization}, utilizing hierarchical memory to respond to user queries by providing relevant context.
   % to the multi-modal language model.
  % Different from nowadays 
  % Vision features saved in the vision buffer are transmitted for memory construction.
  % When a user query comes, it is utilized to search for the most relevant information in our hierarchical memory and feedback to the multi-modal language model as context supplementation.
  }
  \vspace{-4mm}
  \label{figure:overview}
\end{figure*}

\subsection{Hierarchical Memory Storage}\label{hierarchical}
\label{sec:3.1}
$\StreamChat$ treats videos as dynamic information repositories, utilizing hierarchical memory to analyze and store the diverse content.
This section details two specialized memory structures devised to address the challenges of information storage and retrieval: long-short term memory $M_{l} \cup M_{s} = \{l_{i}\}_{i=0}^{T/L} \cup \{s_{i}\}_{i=0}^{S}$  and dialogue memory $M_d = \{d_{i}\}_{i=0}^{D}$. These memories manage visual and conversational data, where $T$ is video duration, $S$ is short memory length, $D$ counts dialogues, and $L$ is the chunk size for long memory. 
The following sections introduce the functions of the above parameters.

\vspace{-2mm}
\subsubsection{Long-short term memory}
\textbf{Selective Frame Stacking.} 
% For the complete video information $V$, 
% \yjz{To reduce the feature storage overhead caused by redundant frames in videos, we use Lucas-Kanade Optical Flow algorithm~\cite{lucas1981iterative} to assist in determining the validity of each video frame $\{ v^i \}_{i=0}^{T} $.}
% Not every frame $\{ v^i \}_{i=0}^{T} $ of the video is effective for analyzing. 
% We use Lucas-Kanade Optical Flow algorithm~\cite{lucas1981iterative} to assist in determining the validity of video frames.
To reduce the feature storage overhead caused by redundant frames in videos, we use Lucas-Kanade Optical Flow algorithm~\cite{lucas1981iterative} in the selective frame stacking module to assist in determining the validity of each video frame $\{ F^i \in \R^{H \times W \times 3} \}_{i=0}^{T} $.
Specifically, we calculate the motion vector $(u, v) $ between $i$-th frames $F^i$ and the last frame $F^{i-1}$:
% \begin{equation}
%    \begin{bmatrix} u \\ v \end{bmatrix} = LKOF(v^m, v^{m-1})
% \end{equation}
\vspace{-1mm}
\setlength{\abovedisplayskip}{2pt} % 调整公式前的间距
\setlength{\belowdisplayskip}{0pt} % 调整公式后的间距
\begin{equation}
\begin{aligned}
    &\quad \begin{bmatrix} u \\ v \end{bmatrix} = 
    \begin{bmatrix}
        \sum_i I_x(i)^2 & \sum_i I_x(i) I_y(i) \\
        \\
        \sum_i I_x(i) I_y(i) & \sum_i I_y(i)^2
    \end{bmatrix}
    ^{-1} 
    \begin{bmatrix}
        \sum_i - I_x(i) I_t(i) \\
        \\
        \sum_i - I_y(i) I_t(i)
    \end{bmatrix} 
    \\
\end{aligned}
\end{equation}
\vspace{-6pt}

where $I_x(i), I_y(i), I_t(i)$ represent the partial derivatives of the frame $F^{i}$ with respect to position $i(x, y)$ and time $t$.
We develop the motion vector magnitude $||\theta|| = \sqrt{u^2 + v^2} \in [0, 1]$ to represent the total motion intensity between frames. 
If $||\theta||$ exceeds the predefined threshold $t \in [0,1]$, the frame $F^i$ will be encoded into vision embedding $e^i \in \R^{n \times d}$ and pushed into buffer
$\mathcal{B}_{\text{vision}}$.
% $\mathcal{B}_{\text{vision}} \leftarrow e^i $. 
% $\mathcal{B}_{\text{vision}} \xleftarrow{\text{push}} e^i $.
% pushed into buffer $\mathcal{B}_{\text{vision}}  $.

% and used to construct basic units in memory. 
% , waiting for memory unit construction. The $\mathcal{B}_{\text{vision}}$ serves as a transfer cache as shown in Fig.~\ref{figure:overview}.
% The length of $\mathcal{B}_{\text{vision}}$ depends on the chunk length during long memory construction.

\begin{figure*}
  \centering
  \setlength{\abovecaptionskip}{0cm}
  \includegraphics[width=1.0\textwidth]{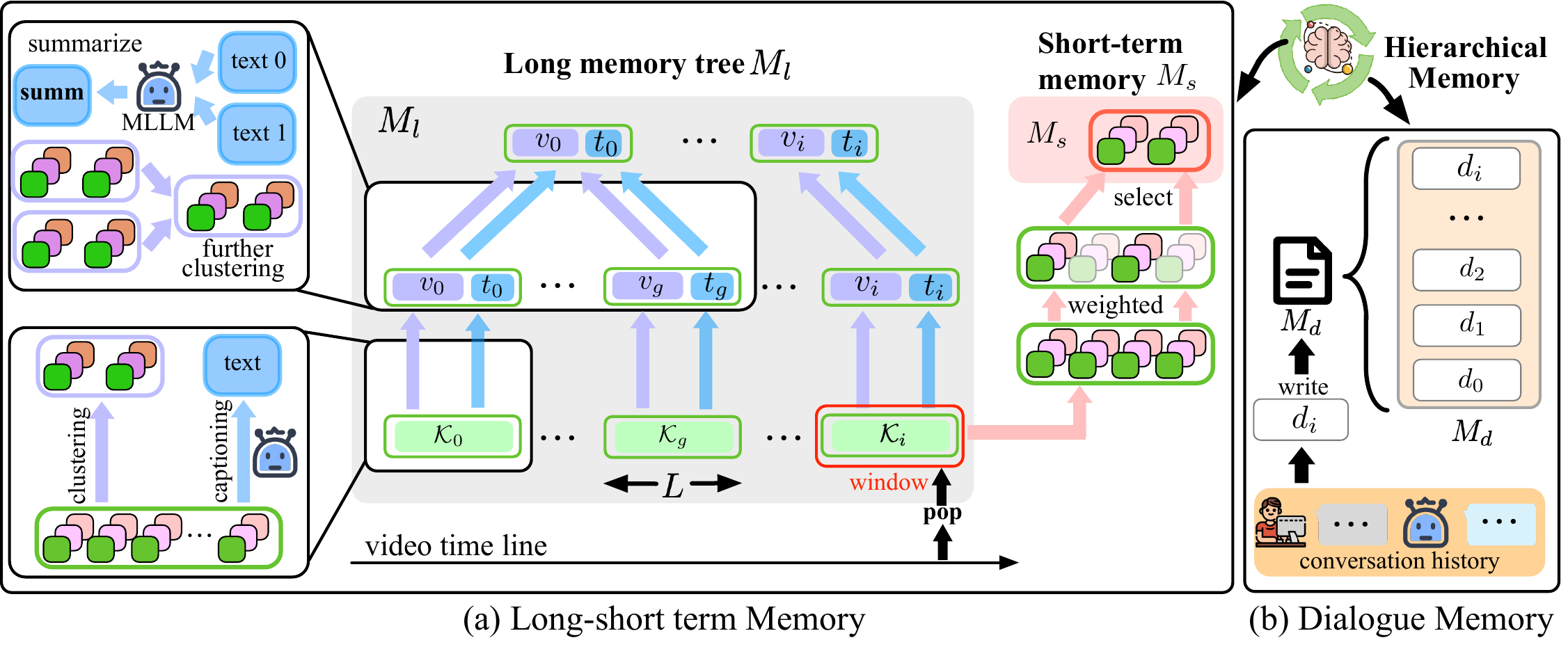}
  % \vspace{-4mm}
  \caption{\textbf{The hierarchical memory storage} (\S\ref{hierarchical}). (a) Long-short term memory, where the long memory tree $M_l$ and short-term memory $M_s$ are constructed along the video time line. (b) The dialogue memory $M_d$ is updated after each inference conversation for managing the dialogue histories.}
  \vspace{-7mm}
  \label{figure:all_memory}
\end{figure*}

\textbf{Short-term Memory.} 
% We argue that short-term memory focuses on preserving clear and specific information over a short period.
% Inspired by the Atkinson-Shiffrin model~\cite{atkinson1968proposed}, which emphasizes the role of short-term memory in preserving clear and specific information for a brief duration with frequent updates, we adapt this concept to manage vision embeddings.
We intend to design a human-like memory method that simulates the Atkinson-Shiffrin model~\cite{atkinson1968proposed}, which emphasizes the role of a short-term storage for maintaining readily accessible, frequently updated information.
Specifically, as shown in Fig.~\ref{figure:all_memory} (a), we select $N$ vision embeddings from the $\mathcal{B}_{\text{vision}}$ as vision candidates $\mathcal{C}$. Building on the Ebbinghaus Forgetting Curve theory~\cite{ebbinghaus2013memory}, we handle memory updates by randomly selecting $S$ vision embeddings $e^i$ from $\mathcal{C}$ to construct the short-term memory $M_s$:
\vspace{-2mm}
\setlength{\abovedisplayskip}{2pt} % 调整公式前的间距
\setlength{\belowdisplayskip}{0pt} % 调整公式后的间距
\begin{equation}
\begin{aligned}
    & \mathcal{C} = \{\sigma_{N-1}e^{i-(N-1)}, \sigma_{N-2}e^{i-(N-2)}, \ldots, \sigma_{0}e^{i}\} \overset{{\tiny\text{random}}}{\underset{{\tiny{\text{select}}}}{\xrightarrow{\hspace{2em}}}} 
    M_s = \{s_{i} \in \R^{n \times d}\}_{i}^{S}
    % M_s = \{s_{0}, s_{1}, \ldots, s_{|S|}\} \subseteq \mathcal{C}, s_{i} = e^i
    % , \quad \sigma_{i} = \frac{e^{-\frac{i}{S}}}{\sum_{j=0}^{N-1} e^{-\frac{j}{S}}}
\end{aligned}
\end{equation}
\vspace{-1mm}
where $\sigma_i$ is the normalized forgetting probability of $i$-th unit of $\mathcal{C}$, $S$ represents the length.

% \subsubsection{long-term memory}
\textbf{Long-term Memory.} 
% \label{sec:3.1}
% Compared to the clear and explicit expression of features in short-term memory, long-term memory is more complex and abstract~\cite{atkinson1968proposed}. 
The long-term memory simulates the complex and abstract memory of humans~\cite{atkinson1968proposed}.
For this reason, we design two forms of information in long-term memory: \textit{text clues}, which is used to store declarative text $t_{i}$ describing events that occurred over a past period, and \textit{vision memory}, which is used to store compressed visual features $v_{i} \in \R^{C\times d}$. 
\textit{Text clues} serve as an index for retrieving relevant information from the long-term memory (introduced in \S\ref{retrieval}).
Our system overcomes the bottleneck of VRAM consumption and the challenges of retrieving memory units $l_{i}$ by constructing a tree structure as shown in Fig.~\ref{figure:all_memory} (a).

The construction of the long-term memory tree can be outlined in the following steps:
\textit{Firstly}, the vision buffer is chunked, and each chunk is clustered and assigned a caption:
% \setlength{\abovedisplayskip}{2pt} % 调整公式前的间距
% \setlength{\belowdisplayskip}{0pt} % 调整公式后的间距
% \begin{equation}
% \begin{aligned}
%     \mathcal{B}_{\text{vision}} &= \{\mathcal{K}_{0}, \mathcal{K}_{1}, \dots, \mathcal{K}_{i}\}, \quad \mathcal{K}_{i} = \{e^{0i}, e^{1i}, \dots, e^{Li}\}, \\
%     v_{|i|m} &= f_{\text{k-means}}(\mathcal{K}_{i}), \quad t_{|i|m} = p_{\theta}(x_i | \mathcal{K}_{i})
% \end{aligned}
% \end{equation}
\vspace{-1mm}
\setlength{\abovedisplayskip}{2pt} % 调整公式前的间距
\setlength{\belowdisplayskip}{0pt} % 调整公式后的间距
\begin{equation}
\begin{aligned}
    % \mathcal{B}_{\text{vision}}=\{\mathcal{K}_{0}, \mathcal{K}_{1}, \dots, \mathcal{K}_{i}\}, \mathcal{K}_{i} = \{e^{0}, e^{1}, \dots, e^{L}\}, v_{|i|}=f_{\text{k-means}}(\mathcal{K}_{i}), t_{|i|} = p_{\theta}(x_i | \mathcal{K}_{i})
    \mathcal{B}_{\text{vision}}=\{\mathcal{K}_{i}\}_{i=0}^{T/L}, \mathcal{K}_{i} = \{e^{i}\}_{i=0}^{L}, v_{i}=f_{\text{k-means}}(\mathcal{K}_{i}), t_{i} = p_{\theta}(x_i | \mathcal{K}_{i})
\end{aligned}
\end{equation}
\vspace{-1mm}
where $v_{i} \in \R^{C\times d}$ is the $i$-th cluster formed from feature chunks 
% \{$\mathcal{K}_{0}, \mathcal{K}_{1}, \dots, \mathcal{K}_{i}$\}
$\{\mathcal{K}_{i}\}_{i=0}^{T/L}$, $C$ is the clustering goals
, and $t_{i}$ represents the $i$-th caption of each chunk.
Each chunk $\mathcal{K}_{i}$ contains $L$ features $e^i \in \R^{n\times d}$ which come from buffer $\mathcal{B}_{\text{vision}}$.
% The chunk length $L$ and clustering $cl$ affect the VRAM space occupied by the tree structure.
\textit{Next}, a clustered feature $v_{i}$ and a caption $t_{i}$ together form a long memory unit $l_{i}$, which also serves as the basis nodes of our tree structure:
\vspace{-1mm}
\begin{equation}
\begin{aligned}
    [ l_{0}, l_{1}, \dots, l_{i-1} ]_{\text{nodes}} \xleftarrow{\text{push}} l_{i} = \{v_{i}, t_{i}\}
\end{aligned}
\end{equation}
\vspace{-5mm}

Finally, basic nodes are further grouped into higher level nodes $[l_{0}^1, l_{1}^1, \dots, l_{k}^1]$ in chronological order until a tree structure is formed and all the basic nodes are exhausted :
\begin{equation}
\begin{aligned}
    % M_l = \{ \overbrace{l_{0m}, l_{1m}}^{l_{0m}^1}, \overbrace{l_{2m}, l_{3m}}^{l_{1m}^1}, \overbrace{l_{4m}, l_{5m}}^{l_{2m}^1}, \dots, \overbrace{l_{|i-1|m}, l_{|i|m}}^{l_{|k|m}^1} \}, \\
    % M_l = \{ l_{0m}, l_{1m}, l_{2m}, l_{3m}, l_{4m}, l_{5m}, \dots, l_{|i-1|m}, l_{|i|m} \}, \\
    M_l = \{l_{i}\}_{i=0}^{T/L},  l_{k}^1 = \{f_{\text{k-means}}(\{v_{i}\}_{i=0}^{g}), p_{\theta}(x_i | \{t_{i}\}_{i=0}^{g}) \}
\end{aligned}
\end{equation}

% \vspace{-3mm}
\subsubsection{Dialogue Memory}
% How we handle multi-round user requests.
% Inspired by ~\cite{zhong2024memorybank}, multi-turn dialogues are also considered a form of memory for iterative updates. 
In our approach, each round of question $\{Q_i\}_{i=0}^{D}$ and answer $\{A_i\}_{i=0}^{D}$ is viewed as a memory fragment, which is pre-encoded by the encoder model  $E(\cdot)$  into a contextual representation $d_{i}$. Thus, the entire dialogue history  $M_{d}$  is pre-encoded as shown in the following formula: $M_d = \{d_{0}, d_{1}, \ldots, d_{i-1}\} \xleftarrow{\text{push}} d_{i} = E(<Q_i, A_i>)$
% $M_{d} = \{d_{0m}, d_{1m}, \ldots, d_{|T|m}\}$ , where each  $d_{|i|m}$  is a vector representation of a memory fragment. These vector representations are then indexed using FAISS~\cite{johnson2019billion} for efficient retrieval. 
% \setlength{\abovedisplayskip}{2pt} % 调整公式前的间距
% \setlength{\belowdisplayskip}{0pt} % 调整公式后的间距
% \begin{equation}
% \begin{aligned}
%     M_d = \{d_{0}, d_{1}, \ldots, d_{i-1}\} \leftarrow d_{i} = E(<Q_i, A_i>)
% \end{aligned}
% \end{equation}
where the length of $M_d$ is equal to the conversation number $D$,
% The user requests $Q$  is encoded by  $E(\cdot)$  into  $h_c$  as a query to search for the most relevant dialogue memory in  $M_d$. 
and we select MiniLM-L6~\cite{wang2020minilm} as our encoder model. 
% because it provides us with accurate retrieval results.
% , which can help us calculate more properly

\vspace{-3mm}
\subsubsection{Retrieval}\label{retrieval}
% When a user question $Q_i$ comes in, the memory system will search for the most relevant knowledge: \textbf{retrieved tokens} ($v_{r}$) and \textbf{context} ($d_{r}$) based on the cosine similarity between the question $Q_i$ and memory units ($t_{i}$ and $d_{i}$).
% In long memory tree $M_l$, the user question $Q_i$ and text memory units $t_{i}$ are encoded by the tokenizer and the embedding layer of LLM. 
% In dialogue memory $M_d$, the user requests $Q_i$  is encoded by  $E(\cdot)$  into  $h_c$  as a query to search for the most relevant memory $d_{|i|}$. 
% After that, the retrieved information is utilized as a context supplement. 
% More details about our retrieval algorithm are shown in$\S$~\ref{sec_app:retrieval_detail}.
When a user question $Q_i$ comes in, the memory system will search for the most relevant knowledge as supplementation by the retrieval algorithm.
In long memory tree $M_l$, the $Q_i$ and text clue units $\{t_{i}\}_{i=0}^{T/L}$ are encoded by the tokenizer and the embedding layer of LLM. Based on the cosine similarity between encoded $Q_i$ and text clue $t_{i}$, the memory system will search for the \textit{retrieved tokens} $M_s \cup \{v_{r} \in \R^{C\times d}\}_{r=0}^{\mathcal{L}}$ where $\mathcal{L}$ represent the layer number of $M_l$.
In dialogue memory $M_d$, the user requests $Q_i$  is encoded by  $E(\cdot)$  as a query to search for the \textit{context} $<Q_{\text{retrieved}}, A_{\text{retrieved}}>$ based on the FAISS~\cite{johnson2019billion} index. 
More details about our retrieval algorithm are shown in Appen.~$\S$\ref{sec_app:retrieval_detail}.

\vspace{-2mm}
% How system schedule mult-process.
\subsection{System Scheduling}
\label{sec:3.2}
As shown in Fig.~\ref{figure:overview}, our method includes three different parts: selective frame stacking, memory formation and contextual summarization.
These components are operated as independent threads to optimize inference speed and minimize latency. System scheduling is crucial as it enables concurrent execution of these threads without interference, significantly enhancing processing speed.
% The reason why the system scheduling makes the agent process video as fast as possible and achieves low latency is that it allows different threads to operate at the same time without interfering with each other.

Specifically, the \textbf{(\romannumeral1)} \textit{slective frame stacking thread} actively populates the vision buffers $\mathcal{B}_{\text{vision}}$ with features $e^i$. Once full, these features are cleared from the buffer and passed to the \textbf{(\romannumeral2)} \textit{memory formation thread}, which updates the memory structures by building the long-term memory tree $M_l$ and refreshing the short-term memory $M_s$.
Concurrently, previous dialogue records ($< Q{i-1}, A_{i-1} >$) are stored in the dialogue memory $M_d$. Upon receiving a new query $Q_i$, the \textbf{(\romannumeral3)} \textit{contextual summarization thread}  
retrieves relevant information from the hierarchical memory to provide timely responses. This architecture supports sub-second latency (<0.9s) and video processing up to 32 FPS.

\section{Experiments}\label{exp}
% To comprehensively validate the superiority of our method, we have tested it in both offline and online scenarios and compared its performance with existing state-of-the-art models. We compared models in the Instruction-tuning category, including Video-LLaVA~\cite{lin2023video}, LLaMA-VID~\cite{li2023llama}, LLaVA-NExT~\cite{liu2024improved}, LongVA~\cite{zhang2024long}, and LLaVA-Hound~\cite{zhang2024direct}, and those in the Training-Free category, such as MovieChat~\cite{song2024moviechat}, FreeVA~\cite{wu2024freeva}, as well as Flash-VStream~\cite{zhang2024flash} and Videollm-online~\cite{chen2024videollm}, which also focus on online video understanding.
% \begin{table}[h]

% This section first introduces the metrics designed to evaluate the performance of streaming video understanding in $\StreamBench$. 
% We then make the model better within the accuracy alignment with different parameters and associations.
% In addition, we further demonstrate that our method can improve the model's performance in offline scenarios.
% We compare our method with various state-of-the-art video understanding multi-modal large language models.
\begin{wraptable}{r}{0.27\textwidth}
% {r}{0.5\textwidth}
    \vspace{-2mm}
    \scriptsize
    \centering
    \caption{\textbf{Memory configurations} for three models.}
    \vspace{-3mm}
    \begin{tabular}{|l|| >{\centering\arraybackslash}p{0.15cm} 
    % >{\centering\arraybackslash}p{0.12cm}
    >{\centering\arraybackslash}p{0.12cm}
    >{\centering\arraybackslash}p{0.12cm}
    >{\centering\arraybackslash}p{0.12cm}|}
        \hline
        % Version & S/$\mathcal{C}$ & $t$ & $L$ & $g$ & C. \\
        \rowcolor{gray!45} Version & $t$ & $L$ & $g$ & $C$ \\
        \hline \hline
        % Slow  & \multirow{3}{*}{5/20} & 0.13  & 35 & 15 & 5      \\
        \texttt{Slow}  &  0.13  & 35 & 15 & 5      \\
        \texttt{Base}   & 0.35             & 25 & 10 & 5       \\
        % \midrule
        \texttt{Fast}   & 0.58             & 30 & 15 &  5    \\
        % \midrule
        \hline
    \end{tabular}
    \vspace{-1mm}
    \label{tab:model_setting}
\end{wraptable}
% \begin{wraptable}{r}{0.27\textwidth}
% % {r}{0.5\textwidth}
%     \vspace{-1mm}
%     \scriptsize
%     \centering
%     \caption{Memory configurations for three models.}
%     \vspace{-3mm}
%     \begin{tabular}{l >{\centering\arraybackslash}p{0.15cm} 
%     % >{\centering\arraybackslash}p{0.12cm}
%     >{\centering\arraybackslash}p{0.12cm}
%     >{\centering\arraybackslash}p{0.12cm}
%     >{\centering\arraybackslash}p{0.12cm}}
%         \toprule
%         % Version & S/$\mathcal{C}$ & $t$ & $L$ & $g$ & C. \\
%         Versions & $t$ & $L$ & $g$ & C. \\
%         \midrule
%         % Slow  & \multirow{3}{*}{5/20} & 0.13  & 35 & 15 & 5      \\
%         Slow  &  0.13  & 35 & 15 & 5      \\
%         Base   & 0.35             & 25 & 10 & 5       \\
%         % \midrule
%         Fast   & 0.58             & 30 & 15 &  5    \\
%         % \midrule
%         \bottomrule
%     \end{tabular}
%     \vspace{-1mm}
%     \label{tab:model_setting}
% \end{wraptable}
\subsection{Experimental Setup}
\textbf{Memory Configurations}. 
To adapt the model to various application scenarios, we configure three versions with different memory settings: Base, Fast, and Slow. These variants adjust key memory parameters, including threshold ($t$), chunk length ($L$), group size ($g$), and clustering goals ($C$), as summarized in Tab.~\ref{tab:model_setting}. The Fast model is optimized for rapid video processing, while the Slow model prioritizes accuracy in responses. The Base model balances processing speed and accuracy.
\vspace{-1mm}
% We introduce three versions of models: Base, Fast and Slow, corresponding to different memory parameters (S: $M_s$ length, $\mathcal{C}$: candidates, $t$: threshold, $L$: chunk length, $g$: group size, C.: clustering goals) as shown in Tab.~\ref{tab:model_setting}. The Fast model is optimized for faster video processing, while the Slow model prioritizes answer correct. The Base model strikes a trade-off between processing speed and accuracy. We compare them with state-of-the-art video understanding multi-modal large language models~\cite{lin2023video, li2023llama, liu2024improved, zhang2024long, zhang2024direct, song2024moviechat, chen2024videollm, zhang2024flash}. All of these versions use CLIP-L-P14~\cite{radford2021learning} with 336 resolution as the vision encoder and utilizes LongVA~\cite{zhang2024long} as our core. All experiments are implemented on 2 NVIDIA Tesla A800 GPUs with 80G memory.

% The specific parameters can be shown in ~\ref{modelsetting}.  
% and all models are based on ~\cite{zhang2024long}.

% \subsection{Metrics}
\textbf{Evaluation Metrics.}
We evaluate semantic similarity in single conversations using the LLaMA-3 model~\cite{dubey2024llama}, which assigns a semantic correctness score (Sco.) ranging from $[0,5]$, where higher scores reflect responses that more closely align with the expected answers. For assessing coherence in multi-turn dialogues, we compute score fluctuations across turns; smaller fluctuations (Coh.) indicate a smoother dialogue experience. Additionally, we measure request processing delay (RPD), defined as the time (in seconds) from user request submission to the start of response generation. A smaller RPD signifies lower latency, resulting in reduced wait times for users. Appen.~$\S$\ref{sec_app:metrics_detail} offers more details.
% We use a language model to assess the semantic similarity of single conversations. Our benchmark uses LLaMA-3~\cite{dubey2024llama} as our scoring model to generate a semantic correctness score from $[0,5]$. 
% Higher scores indicate that the model's response is closer to the answer.
% \vspace{-1mm}

% \textbf{Coherence:}
% We calculated the score fluctuations between multi-turn dialogues to measure coherence. Smaller coherence indicates that the model provides a smoother dialogue experience for users.
% \vspace{-1mm}

% \textbf{Request Processing Delay:}
% RPD measures the time interval (in seconds) from when a user submits a request to when the response generation begins. A smaller RPD indicates reduced latency, meaning users experience shorter wait times before receiving answers.

\textbf{Implementation Details.}
We utilize CLIP-L-P14~\cite{radford2021learning} as the vision encoder and we set the number of selected memory units $S$ to 5 and candidate length $\mathcal{C}$ to 20. Experiments were conducted on two NVIDIA Tesla A800 GPUs with 80GB of memory each (more details in Appen.~$\S$\ref{sec_app:selection} ). We benchmark our model against state-of-the-art methods, including Video-LLaVA~\cite{lin2023video}, LLaMA-VID~\cite{li2023llama} and etc.
% \cite{lin2023video, li2023llama, liu2024improved, zhang2024long, zhang2024direct, song2024moviechat, chen2024videollm, zhang2024flash}.
% To verify whether the model can generate responses in the shortest theoretical time, we use user request processing delay (RPD) as an indicator which measures the time difference from when the user sends a request to when the response generation begins (measured in seconds). Smaller RPD indicates less latency meaning users need to wait a little time to get answers. 
% To test the RDP latency of offline methods, we defined it as the time from video processing to the start of generation, which includes video sampling and encoding as benchmarks for evaluation.

\subsection{Comparison with State-of-the-art Methods}
\begin{wraptable}{r}{0.53\textwidth}
    \scriptsize
    \centering
    \vspace{-1mm}
    \caption{\textbf{Quantitative results in StreamBench}. RPD is measured for streaming methods. Fr.: sampled frames.}
    % The best and second methods are denoted with \textbf{bold} and \underline{underline}. 
    % F.means Frames/FPS.
    \vspace{-3mm}
    \begin{tabular}{|l||>{\centering\arraybackslash}p{0.32cm} | >{\centering\arraybackslash}p{0.32cm} |
    >{\centering\arraybackslash}p{0.32cm}
    >{\centering\arraybackslash}p{0.32cm}
    >{\centering\arraybackslash}p{0.32cm}
    >{\centering\arraybackslash}p{0.32cm}|}
    % \begin{tabular}{lccccc}
        \hline
        \rowcolor{gray!45} ~~~~~~~~~ Method        & FPS & Fr. & Sco. & Acc. & Coh. & RPD\\
        \hline \hline
        ~  {Human performance}  &  -  & -                  &   {4.03}  &    {79.4}  &    {1.16} & -   \\
        ~  {GPT-4o~\cite{hurst2024gpt}}  &  -  &   {50}                  &   {3.70}  &    {71.0}  &    {1.66}      & -                \\
        ~  {GPT-4o~\cite{hurst2024gpt}}  &  -  &   {35}                  &   {3.64}  &    {69.8}  &    {1.72}     & -          \\
        ~  {GPT-4o-mini~\cite{hurst2024gpt}}  &  -  &   {35}             &   {3.17} &   {59.1} &   {2.01}       & -                \\
        \rowcolor{gray!10}
        \multicolumn{7}{|l|}{\textit{Instruct-tuning}} \\
        ~Video-LLaVA~\cite{lin2023video}  &  -  & 8                  & 2.81           & 48.9              & 2.19             & -                \\
        ~LLaMA-VID~\cite{li2023llama}    &  -    & 180               & 2.94           & 51.2              & 2.08            & -                \\
        ~LLaVA-NExT~\cite{liu2024llavanext}   &  -    & 8              & 2.65            & 46.2              & 2.18            & -                 \\
        ~LLaVA-Hound~\cite{zhang2024direct}   &  -   &   8          & 3.12            & 54.7              & 1.83             & -       \\
        ~LongVA~\cite{zhang2024long}     &  -      & 8                & 3.05            & 52.4              & 1.96             & -                \\
        ~  {MiniCMP-v2.6~\cite{yao2024minicpm}}     &  -      &   {8}       &   {2.97}            &   {56.6}              &   {2.21}             & -                \\
        ~  {VILA1.5~\cite{lin2024vila}}     &  -      &   {8}                &   {3.10}            &   {57.1}              &   {2.20}            & -                \\
        ~  {InternVL2~\cite{chen2024internvl}}   &  -   &   {8}                &   {3.15}            &   {57.6}              &   {2.11}             & -                \\
        ~  {InternLM-XCP2.5~\cite{zhang2024internlm}}     &  -      &   {8}                &   {3.21}            &   {57.7}              &   {2.12}             & -                \\
        \hline
        \rowcolor{gray!10}
        \multicolumn{7}{|l|}{\textit{Training-free}} \\
        ~MovieChat~\cite{song2024moviechat}   &  -    & 32               & 2.07            & 35.3             & 2.36             & -                \\
        ~FreeVA ~\cite{wu2024freeva}     &  -    & 4             & 3.10           & 56.3              & 2.11             & -                \\
        \hline
        \rowcolor{gray!10}
        \multicolumn{7}{|l|}{\textit{Streaming}} \\
        % \hline
        ~Video-online~\cite{chen2024videollm}    & 5     &  -        & 3.11              & 56.4              & 1.94	         & 1.07               \\
        ~Flash-VStream~\cite{zhang2024flash}   & 1    &  -         & 2.89              & 52.1              & 2.21	         & 4.15               \\
        \hline
        \rowcolor{gray!10}
        \multicolumn{7}{|l|}{\StreamChats} \\
        ~~~\texttt{Slow}  & 15      &  -       & \textbf{3.48}  & \textbf{64.7}	 & \textbf{1.76}	 &{0.90} \\
        % \rowcolor{cyan!20}
        ~~~\texttt{Base}   & \underline{20}  &  -  &  \underline{3.42}  &  \underline{63.8} &  \underline{1.79} & \underline{0.89}  \\
        ~~~\texttt{Fast}   & \textbf{32}   &  -  & 3.28  & 61.7  & 1.81 & \textbf{0.85} \\
        \hline
    \end{tabular}
    \vspace{-3mm}
    \label{tab:model_comparison}
\end{wraptable}

\textbf{Online Scenarios.}  
As shown in Tab.~\ref{tab:model_comparison}, our models demonstrate significant improvements over the previous best method, Video-online~\cite{chen2024videollm}.
\begin{itemize}[leftmargin=*,itemsep=0pt]\vspace{-2mm}
    \item \texttt{Slow}: Achieves an \textbf{8.3\%} higher accuracy and a \textbf{0.37} higher score than Video-online.
    \item \texttt{Fast}: Processes video at 32 FPS, making it much faster than all previous streaming methods, while still improving accuracy by \textbf{5.3\%} and scoring \textbf{0.17} higher than~\citep{chen2024videollm}.
    \item \texttt{Base}: Reaches 63.8\% Acc. and 3.42 score.
    \item Best model: Surpasses ~\citep{chen2024videollm} with a \textbf{0.18} improvement in coherence score and reduces latency by \textbf{0.17s}, delivering smoother conversations with shorter wait times.
    \vspace{-2mm}
\end{itemize}

Due to system scheduling, all models maintain nearly the same response time of about 0.9s.  
Tab.~\ref{tab:types_score} presents the detailed scores across six question types. Using hierarchical memory storage, our method excels in object search (OS), long-term memory search (LM), short-term memory search (SM), and conversational interaction (CI) tasks.
Notably, our \texttt{Slow} model increases accuracy by \textbf{10.3\%} in OS, \textbf{5.1\%} in LM, \textbf{4.9\%} in SM, and \textbf{5.8\%} in CI compared to Video-online.

\begin{table*}[t]
\scriptsize
\centering
\label{tab:model_evaluation}
\vspace{-10pt}
\caption{\textbf{Quantitative comparison across six tasks.} Detailed results for tasks `OS', `LM', `SM', `CI', `KG', and `SF'. For full names and definitions, refer to $\S$\ref{sec2.2}}.
% We label the best and second methods with \textbf{bold} and \underline{underline} styles.}
\vspace{-10pt}
\begin{tabular}{|l|l||
    >{\centering\arraybackslash}p{0.3cm} 
    >{\centering\arraybackslash}p{0.4cm}| 
    >{\centering\arraybackslash}p{0.3cm}
    >{\centering\arraybackslash}p{0.4cm}|
    >{\centering\arraybackslash}p{0.3cm}
    >{\centering\arraybackslash}p{0.4cm}| 
    >{\centering\arraybackslash}p{0.3cm}
    >{\centering\arraybackslash}p{0.4cm}|
    >{\centering\arraybackslash}p{0.3cm}
    >{\centering\arraybackslash}p{0.4cm}|
    >{\centering\arraybackslash}p{0.3cm}
    >{\centering\arraybackslash}p{0.4cm}|}
\hline
% \rowcolor[HTML]{EFEFEF} 
\rowcolor{gray!45} ~ & ~ & \multicolumn{2}{c|}{OS} & \multicolumn{2}{c|}{LM} &  \multicolumn{2}{c|}{SM} &  \multicolumn{2}{c|}{CI} &  \multicolumn{2}{c|}{KG} & \multicolumn{2}{c|}{SF}\\ 
\rowcolor{gray!45} ~~~~~~~~ Method & ~~~Publication & \textit{Sco.} & \textit{Acc.} & \textit{Sco.} & \textit{Acc.} & \textit{Sco.} & \textit{Acc.} & \textit{Sco.} & \textit{Acc.} & \textit{Sco.} & \textit{Acc.} & \textit{Sco.} & \textit{Acc.} \\
\hline
\hline
~  {Human performance} & ~~~~~~~~~ - - &   {3.95} &   {71.8} &   {3.81} &   {69.3} &   {4.07} &   {81.5} &   {4.14} &   {82.6} &   {4.06} &   {80.7} &   {4.30} &   {80.7} \\
~{GPT-4o-50~\citep{hurst2024gpt}} &  {Arxiv~~~~~2024} &   {3.27} &   {60.5} &   {3.35} &   {61.2} &   {3.41} &   {64.4} &   {3.81} &   {72.3} &   {4.58} &   {93.9} &   {3.83} &   {74.7} \\
~{GPT-4o-35~\citep{hurst2024gpt}} &   {Arxiv~~~~~2024} &   {3.22} &   {59.6} &   {3.28} &   {58.6} &   {3.45} &   {65.3} &   {3.76} &   {71.7} &   {4.54} &   {93.3} &   {3.50} &   {66.1}   \\
~{GPT-4o-mini-35~\citep{hurst2024gpt}} &   {Arxiv~~~~~2024} &   {2.52} &   {46.8} &   {2.70} &   {45.8} &   {2.80} &   {51.0} &   {3.50} &   {64.0} &   {4.67} &   {95.2} &   {2.90} &   {53.3}  \\
\hline
\rowcolor{gray!10}
\multicolumn{14}{|l|}{\textit{Instruct-tuning}} \\

~Video-LLaVA~\citep{lin2023video} & EMNLP~2024 & 2.25 & 31.2 & 2.31 & 35.9 & 2.50 & 41.8 & 3.18 & 56.1 & 3.81 & 74.6 & 2.93 & 54.8 \\
~LLaMA-VID~\citep{li2023llama}   & ECCV~~~~2024 & 2.32 & 33.9 & 2.43 & 38.2 & 2.63 & 44.1 & 3.31 & 58.4 & 3.93 & 76.9 & 3.06 & 57.1 \\
~  {VILA1.5~\citep{lin2024vila}}      &   {CVPR~~~~2024} &   {2.33} &   {36.1} &   {2.54} &   {44.3} &   {2.87} &   {50.8} &   {3.59} &   {68.3} &   {3.97} &   {78.6} &   {3.38} &   {65.5}  \\
~  {InternVL2~\citep{chen2024internvl}}      &   {CVPR~~~~2024} &   {2.49} &   {38.5} &   {2.70} &   {46.6} &   {2.89} &   {50.9} &   {3.61} &   {67.6} &   {4.02} &   {81.0} &   {3.29} &   {62.2}  \\
~LLaVA-NExT~\citep{liu2024llavanext}  & Arxiv~~~~~2024 & 2.17 & 35.0 & 2.14 & 31.4 & 2.15 & 36.0 & 2.55 & 42.7 & 3.88 & 76.1 & 3.12 & 57.6 \\
~LLaVA-Hound~\citep{zhang2024direct} & Arxiv~~~~~2024 & 2.49 & 37.6 & 2.68 & 43.2 & 3.09 & 53.4 & 3.21 & 55.7 & 3.89 & 76.3 & 3.35 & 62.0 \\ 
~LongVA~\citep{zhang2024long}      & Arxiv~~~~~2024 & 2.61 & 41.8 & 2.81 & 47.4 & \underline{3.20} & \underline{57.6} & 3.29 & 59.8 & 4.01 & 80.7 & 3.48 & 66.1 \\
~  {MiniCMP-v2.6~\citep{yao2024minicpm}}      &   {Arxiv~~~~~2024} &   {2.32} &   {37.6} &   {2.78} &   {51.9} &   {2.62} &   {43.7} &   {3.35} &   {65.7} &   {3.19} &   {66.2} &   {3.27} &   {64.2} \\
~  {InternLM-XCP2.5~\citep{zhang2024internlm}}      &   {Arxiv~~~~~2024} &   {2.40} &   {38.8} &   {2.81} &   {43.3} &   {2.89} &   {50.8} &   {3.62} &   {65.6} &   {4.41} &   {88.4} &   {3.23} &   {60.5} \\
\hline
        \rowcolor{gray!10}
\multicolumn{14}{|l|}{\textit{Training-Free}} \\

% \multirow{2}{*}{Training-Free} 
~MovieChat~\citep{song2024moviechat}  & CVPR~~~~2024 & 1.45 & 18.6 & 1.42 & 20.4 & 1.76 & 26.5 & 2.28 & 42.3 & 3.39 & 67.2 & 2.05 & 35.8 \\
~FreeVA~\citep{wu2024freeva}    & Arxiv~~~~~2024 & 2.39 & 35.6 & 2.33 & 37.5 & 2.62 & 43.7 & 3.16 & 58.8 & 4.24 & 84.0 & 2.87 & 53.7 \\ 
\hline
        \rowcolor{gray!10}
\multicolumn{14}{|l|}{\textit{Online}} \\

~Video-online~\citep{chen2024videollm} & CVPR~~~~2024 & 2.61 & 41.4 & \underline{2.87} & 48.8 & 3.01 & 52.9 & 3.31 & 62.7 & 3.58 & 69.2 & 3.39 & 64.1 \\
~Flash-VStream~\citep{zhang2024flash}   & Arxiv~~~~~2024 & 2.38 & 37.1 & 2.64 & 44.5 & 2.78 & 48.6 & 3.13 & 58.1 & 3.34 & 66.4 & 3.17 & 59.2 \\
% \midrule
% \rowcolor{cyan!20}
\hline
\rowcolor{gray!7}
\multicolumn{14}{|l|}{\StreamChats} \\
~~~\texttt{Slow} & ~~~~~~~~~ - - & \textbf{3.01} & \textbf{51.7} & \textbf{2.93} & \textbf{53.9} & \textbf{3.21} & \textbf{57.8} & \textbf{3.86} & \textbf{68.5} & \textbf{4.38} & \textbf{88.1} & \textbf{3.57} & \textbf{69.3} \\
~~~\texttt{Base} & ~~~~~~~~~ - - & \underline{2.93} & \underline{50.5} & \underline{2.87} & \underline{52.9} & 3.15 & 56.1 & \underline{3.82} & \underline{67.6} & \underline{4.37} & \underline{87.9} & \underline{3.56} & \underline{68.8} \\ 
~~~\texttt{Fast} & ~~~~~~~~~ - - & 2.78 & 48.1 & 2.73 & 49.5 & 3.02 & 53.5 & 3.69 & 65.2 & 4.12 & 86.7 & 3.46 & 67.6 \\ 
\hline
\end{tabular}
\label{tab:types_score}
\end{table*}
% \vspace{-16pt}

\textbf{Offline Scenarios.}
We compare our \texttt{Base} model against other methods in general offline video understanding benchmarks including MSRVTT-QA~\citep{xu2016msr}, ActivityNet~\citep{yu2019activitynet}, NExT-QA~\citep{xiao2021next}, MSVD-QA~\citep{xu2017video}.
Since these benchmarks involve open-ended questions, 
we evaluate performances using score and accuracy as metrics, employing the same score model~\cite{dubey2024llama} as used in online tests.
It should be noted that considering the limited average video length in the MSRVTT~\cite{xu2016msr} and MSVD~\cite{xu2017video} shown in Table~\ref{tab:benchmark_comparison}, we did not apply long-term memory $M_l$ for our model during the test. 
Additionally, since these open-ended question-answering test format benchmarks do not evaluate multi-round dialogue capabilities, we removed the dialogue memory $M_d$ component from our model.
\begin{itemize}[leftmargin=*,itemsep=0pt]\vspace{-2mm}
    \item In the MSVD and MSRVTT benchmarks, 
    % our short-term memory provides more specific vision information, 
    our short-term memory module allows the model to capture more specific visual details,
    leading to accuracies of 58.7\% and 43.4\%, respectively.
    % , which is 0.9\% and 1.0\% higher than the previous methods.
    \item With the integration of long-term memory module, our model enhances the longer video performance, surpassing the previous best streaming method Flash-VStream~\citep{zhang2024flash} for \textbf{12.8\%} and the best offline method LLaVA-Hound~\citep{zhang2024direct} for \textbf{1.4\%} in ActivityNet~\citep{yu2019activitynet} benchmark.
    In NExT-QA~\cite{xiao2021next},   {our method can further improve the foundation model LongVA~\citep{zhang2024long} by \textbf{5.1\%} in accuracy}.
    \item   {Although the base model LongVA~\citep{zhang2024long} has achieved the best average accuracy in offline benchmarks}, our method further improves it by \textbf{2.5\%}, proving the effectiveness of the memory module.
    \vspace{-2mm}
\end{itemize}

\begin{table}
% \footnotesize
\centering
% \begin{minipage}{0.37\textwidth}
    % \fontsize{8.3pt}{9pt}\selectfont
    \scriptsize
    \setlength{\tabcolsep}{1.0pt}
    % \vspace{-8pt}
    \caption{\textbf{Performance comparison} of various models in offline video understanding benchmark.}
    \vspace{-8pt}
    \begin{tabular}{|l|l|| 
    >{\centering\arraybackslash}p{0.9cm}
    >{\centering\arraybackslash}p{0.9cm}|
    >{\centering\arraybackslash}p{0.9cm}
    >{\centering\arraybackslash}p{0.9cm}|
    >{\centering\arraybackslash}p{0.9cm}
    >{\centering\arraybackslash}p{0.9cm}|
    >{\centering\arraybackslash}p{0.9cm}
    >{\centering\arraybackslash}p{0.9cm}|
    >{\centering\arraybackslash}p{0.9cm}
    >{\centering\arraybackslash}p{0.9cm}|}
        \hline
        \rowcolor{gray!45} ~ & ~ & \multicolumn{2}{c|}{ActNet} & \multicolumn{2}{c|}{NExT-QA} & \multicolumn{2}{c|}{MSVD} & \multicolumn{2}{c|}{MSRVTT} & \multicolumn{2}{c|}{Average} \\
        \cline{3-12}
        % \addlinespace[1.5pt]

        \rowcolor{gray!45} ~~~~~~~~~ Method & ~~~Publication & \textit{Sco.} & \textit{Acc.} & \textit{Sco.} & \textit{Acc.} & \textit{Sco.} & \textit{Acc.} & \textit{Sco.} & \textit{Acc.} & \textit{Sco.} & \textit{Acc.} \\
            \hline
        \hline
        ~Video-LLaVA~\citep{lin2023video}   & ~EMNLP~2024~ & 1.96 & 35.8 & 2.02 & 34.9 & 2.94 & 57.5 & 2.24 & 42.8 & 2.29 & 42.7           \\
        ~LLaMA-VID~\citep{li2023llama}     & ~ECCV~~~~2024~ & 2.09 & 36.6 & 2.07 & 36.0 & 2.83 & 56.9 & 2.23 & 42.6 & 2.30 & 43.1         \\
        ~MovieChat~\citep{song2024moviechat}     & ~CVPR~~~~2024~ & 2.27 & 37.8 & 2.05 & 35.6 & 2.97 & \underline{57.9} & 2.15 & \underline{43.0} & 2.36 & 43.5         \\
        ~Video-online~\citep{chen2024videollm}  & ~CVPR~~~~2024~ & 2.01 & 36.5 & 2.03 & 35.8 & 2.87 & 54.2 & 2.06 & 38.2 & 2.24 & 41.1          \\
        % LLaVA-NeXT-8B    & - & - & 2.75	& 54.3 & 2.19 & 41.2                 \\
        ~LongVA~\citep{zhang2024long}        & ~Arxiv~~~~~2024~ & 2.48 & 47.1 & \underline{2.74} & \underline{45.4} & 2.98 & 57.8 & 2.22 & 42.4 & 2.60 & \underline{48.1}         \\
        ~LLaVA-Hound~\citep{zhang2024direct}~~   & ~Arxiv~~~~~2024~ & \underline{2.69} & \underline{48.7} & 2.56 & 43.7 & \underline{3.07} & 56.8 & \textbf{2.42} & 42.7 & \underline{2.68} & 47.9         \\
        ~FreeVA~\citep{wu2024freeva}        & ~Arxiv~~~~~2024~ & 2.48 & 46.7 & 2.32 & 41.7 & 3.02 & 58.1 & 2.16 & 38.3 & 2.49 & 46.2        \\
        ~Flash-VStream~\citep{zhang2024flash} & ~Arxiv~~~~~2024~ & 2.02 & 37.3 & 2.06 & 36.1 & 2.91 & 56.1 & 2.08 & 39.8 & 2.26 & 42.3          \\
        \hline
        % \rowcolor{cyan!20}
        % \textbf{LongVA with memory} & - & - & - & - & - & - \\
        ~\StreamChats          & ~~~~~~~~~ - - & \textbf{2.78} & \textbf{50.1} & \textbf{2.84} & \textbf{50.5} & \textbf{3.08}	& \textbf{58.7} & \underline{2.38} & \textbf{43.4} & \textbf{2.77} & \textbf{50.6}  \\
        \hline
    \end{tabular}
    \vspace{-12pt}
    \label{tab:off_line}
\end{table}

\vspace{-5mm}
\subsection{Case Study}\label{case_study}
\begin{figure*}[h]
  \centering
  % \vspace{-4mm}
  \setlength{\abovecaptionskip}{0cm}
  \includegraphics[width=1.0\textwidth]{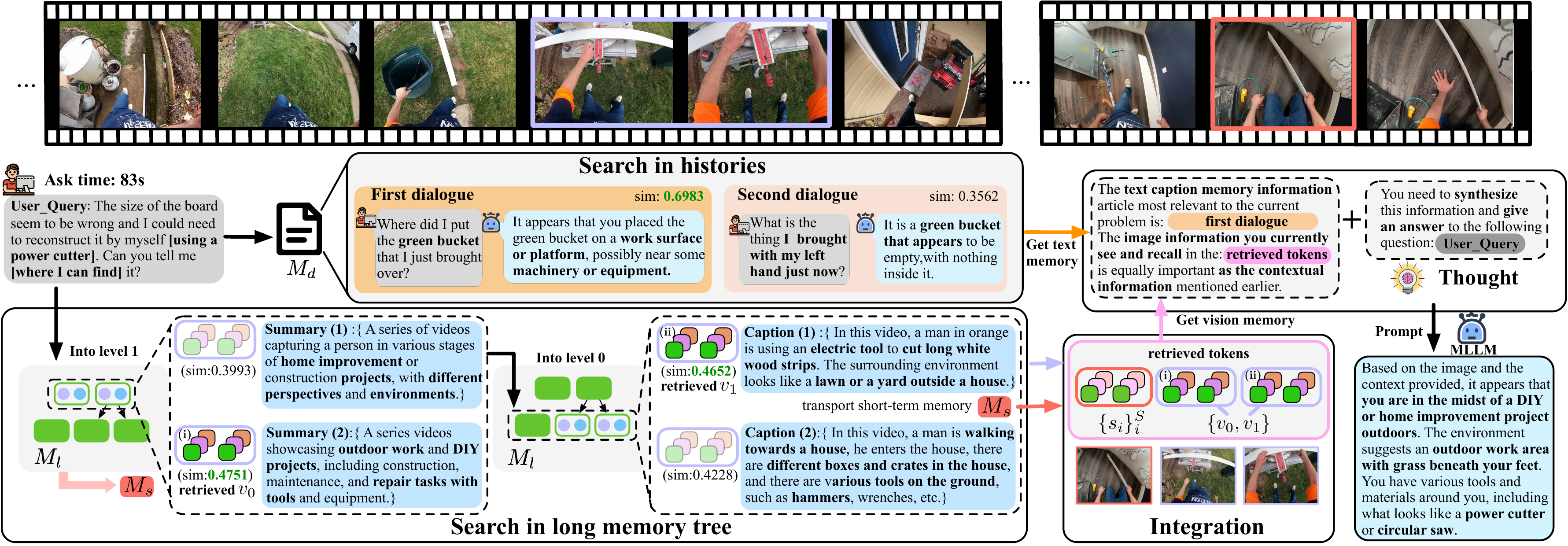}
  \vspace{-3mm}
  \caption{
  \textbf{An inference example of StreamChat} (\S\ref{case_study}).
  Given a question, our system retrieves the most related information in a long memory tree and dialogue histories based on the highest cosine similarity.
  % After that, it integrates retrieved information as a prompt to generate the correct answer.
  % This figure shows the specific content and integrated format of the intermediate steps during searching.
  }
  \vspace{-3mm}
  \label{figure:case_study}
\end{figure*}

In Fig.~\ref{figure:case_study}, we illustrate the reasoning process of $\StreamChat$ with $g=2$ to simplify the observation of internal mechanisms.  The scenario involves a user asking $\StreamChat$ to identify a tool meeting specific requirements and to describe its environment. The memory structure consists of a dialogue memory, $M_d$, with two historical entries, and a layered memory, $M_l$, with two levels. The memory tree visualization shows that the system initially searches for key information at level 1. It computes cosine similarity between the user's query $Q_i$ and two memory units, Summary (1) and (2), obtaining scores of 0.3993 and 0.4751, respectively. Based on these results, $\StreamChat$ selects the path from the second node ($v_1$) due to its higher similarity score and continues along this path. Subsequently, the system aggregates value $\{v_{r}\}_{r=0}^{1}$ from $M_s$ into retrieved tokens that are then incorporated into the reasoning process. Additionally, a high similarity score of 0.6983 between $Q_i$ and the first historical conversation helps provide context, enhancing the depth and relevance of the response.
% As shown in the memory tree, the system first performs a key information search at level 1. By calculating the cosine similarity between the user query $Q_i$ and two units $\{t_{i}\}_{i=0}^{1}$, Summary (1) and (2), with results of 0.3993 and 0.4751 respectively, $\StreamChat$ selects the $v_{1}$ from the second node and continues the search along this path.
% The system integrates the $\{v_{r}\}_{r=0}^{1}$ and $M_s$, combining them into retrieved tokens and feeding to the thought process.
% Additionally, the system identifies a high similarity (0.6983) between the $Q_i$ and the first conversation, provided as the related context for reasoning.

\subsection{Ablation Study}\label{Ablation}
\setlength{\tabcolsep}{1.5pt}
\begin{table}[t]
\centering
\scriptsize
\label{tab:ablation}
\caption{\textbf{Analysis of hierarchical memory.} This table shows the impact of various memory configurations.}
\vspace{-10pt}
\begin{tabular}{|>{\centering\arraybackslash}p{0.36cm} >{\centering\arraybackslash}p{0.36cm} 
    >{\centering\arraybackslash}p{0.36cm}||
    >{\centering\arraybackslash}p{0.7cm}
    >{\centering\arraybackslash}p{0.7cm}|
    >{\centering\arraybackslash}p{0.7cm}
    >{\centering\arraybackslash}p{0.7cm}|
    >{\centering\arraybackslash}p{0.7cm}
    >{\centering\arraybackslash}p{0.7cm}|
    >{\centering\arraybackslash}p{0.7cm}
    >{\centering\arraybackslash}p{0.7cm}|
    >{\centering\arraybackslash}p{0.7cm}
    >{\centering\arraybackslash}p{0.7cm}|
    >{\centering\arraybackslash}p{0.7cm}
    >{\centering\arraybackslash}p{0.7cm}|
    >{\centering\arraybackslash}p{0.7cm}
    >{\centering\arraybackslash}p{0.7cm}|}
% \begin{tabular}{|c|c|c||c|c|c|c|c|c|c|c|c|c|c|c|c|c|}

\hline
% \rowcolor[HTML]{EFEFEF} 
% \multirow{2}{*}{\textbf{$M_l$}} & \multirow{2}{*}{\textbf{$M_s$}} \multirow{2}{*}{\textbf{$M_d$}}
\rowcolor{gray!45} ~ & ~ & ~  & \multicolumn{2}{c|}{OS} & \multicolumn{2}{c|}{LM} & \multicolumn{2}{c|}{SM} & \multicolumn{2}{c|}{CI} & \multicolumn{2}{c|}{KG} & \multicolumn{2}{c|}{SS}  & \multicolumn{2}{c|}{Average} \\ 
% \cline{4-17}
% \cmidrule(lr){4-15}
\rowcolor{gray!45} $M_l$ & $M_s$ & $M_d$ &  \textit{Sco.} & \textit{Acc.} & \textit{Sco.} & \textit{Acc.} & \textit{Sco.} & \textit{Acc.} & \textit{Sco.} & \textit{Acc.} & \textit{Sco.} & \textit{Acc.} & \textit{Sco.} & \textit{Acc.} & \textit{Sco.} & \textit{Acc.} \\ 
\hline \hline
% \multirow{5}{*}{Instruct-tuning}
\xmark & \xmark & \xmark & 2.54 & 41.6 & 2.55 & 45.5 & 2.93 & 52.5 & 3.30 & 60.1 & \textbf{4.44} & \textbf{89.9} & \textbf{3.79} & \textbf{72.6} & 3.27 & 60.3 \\ 
\xmark & \xmark & \cmark & 2.55 & 41.9 & 2.55 & 45.7 & 2.94 & 52.5 & \underline{3.66} & \underline{64.2} & \textbf{4.44} & \underline{88.7} & \underline{3.78} & \underline{72.4} & 3.32 & 60.9 \\
\xmark & \cmark & \xmark & 2.58 & 43.3 & 2.62 & 46.6 & 3.09 & 55.7 & 3.31 & 60.7 & 4.39 & 88.1 & 3.68 & 69.8 & 3.28 & 60.7 \\ 
\cmark & \xmark & \xmark & 2.85 & 49.5 & 2.78 & 51.7 & 2.96 & 53.5 & 3.32 & 61.1 & \underline{4.42} & 88.4 & 3.65 & 69.4 & 3.33 & 62.2 \\ 
\cmark & \cmark & \xmark & \underline{2.91} & \underline{50.4} & \textbf{2.88} & \textbf{53.0} & \underline{3.10} & \underline{56.0} & 3.55 & 63.4 & 4.36 & 87.6 & 3.58 & 68.7 & \underline{3.39} & \underline{63.1} \\ 
\cmark & \cmark & \cmark & \textbf{2.93} & \textbf{50.5} & \textit{2.87} & \textbf{52.9} & \textbf{3.15} & \textbf{56.1} & \textbf{3.82} & \textbf{67.6} & 4.37 & 87.9 & 3.56 & 68.8 & \textbf{3.42} & \textbf{63.8} \\ 
\hline 
\end{tabular}
\vspace{-16pt}
\label{tab:ablation}
\end{table}

\textbf{Exploring Effects of Hierarchical Memory.}
We conduct ablation experiments using the \texttt{Base} model to assess the impact of different memory components on performance. As shown in Table~\ref{tab:ablation}, adding $M_d$ to the base model improved performance on the CI task by 4.1\% without affecting other tasks.
Adding $M_l$ improved the LM task performance by 6.2\%, while the use of $M_s$ boosts SM task performance by 3.2\%.
The results indicate that the model’s performance in each subtask aligns with the inclusion of specific memory attributes.
Additionally, we observe that different memory components can complement each other. When both long-term $M_l$ and short-term memory $M_s$ are applied simultaneously, the average accuracy increases by 0.9\%.

\begin{figure}[bth]
  \centering
  \setlength{\abovecaptionskip}{0cm}
  \includegraphics[width=1.0\textwidth]{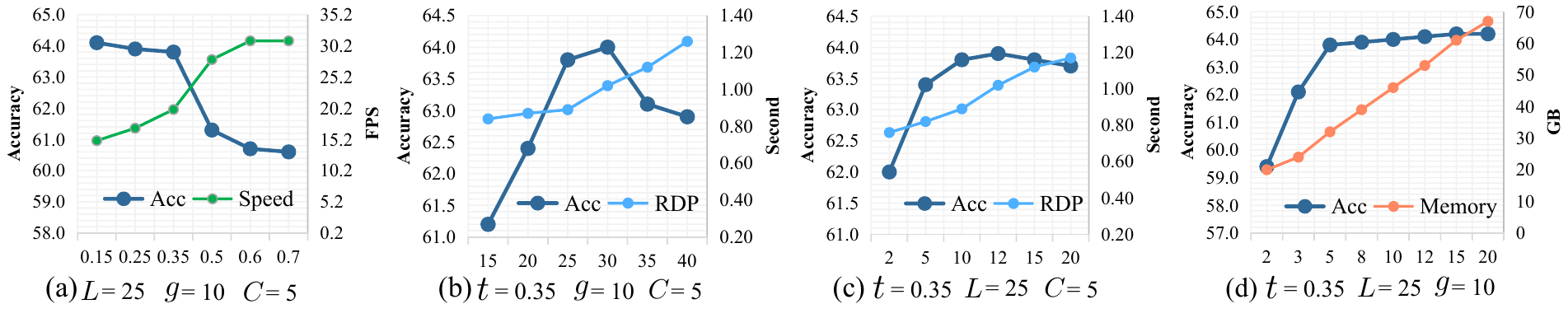}
  % \vspace{-4mm}
  % \vspace{2mm}
  \caption{
  \textbf{Analysis of memory parameters}. (a) The influence between speed and threshold; Impact of (b) chunk length and (c) group size on performance and latency; (d) Effect of clustering goal on performance and VRAM.
  }
  \vspace{-4mm}
  \label{figure:lgc}
\end{figure}

\textbf{Tradeoffs in Speed and Threshold Settings.} 
% The manually set threshold of the Lucas-Kanade Optical Flow algorithm can directly affect the speed of video processing. 
% We record the relationship between setting different thresholds and video process speed in Fig.~\ref{figure:lgc} (a).
% It can be observed that the video processing speed is negatively correlated with the threshold. The bigger the threshold, the faster the actual processing speed.
% But the increase is not limited, when the threshold $t$ reaches $0.55$ the speed no longer increases.
The threshold of the Lucas-Kanade Optical Flow algorithm significantly influences video processing speeds.  As illustrated in Fig.~\ref{figure:lgc} (a),  increasing the threshold initially accelerates the processing speed. However, this increase saturates when $t$ reaches 0.55, stabilizing at 32 FPS.
% In addition, using too high a video processing speed is not recommended, because we observe that the video processing speed can affect the model performance by reducing the available video frames. 
% A higher threshold implies a greater requirement for the degree of change in adjacent video frames.
% The original video frames that can be saved will be lost because the threshold is set too high, so the original video information that can be used in the entire memory storage will be significantly reduced.
Importantly, higher processing speeds are discouraged due to their detrimental impact on model performance (64.0\%$\rightarrow$60.7\%). 
Elevating thresholds leads to more pronounced changes in frame differences and loss of original data, thereby limiting the model's ability to effectively utilize the full spectrum of video information.
% Consequently, the reduction in available video information compromises the model’s ability to effectively process and utilize the complete range of video data.

\textbf{Design of Long Memory Tree.} 
The chunk length ($L$), group size ($g$), and clustering goal ($C$) significantly impact the effectiveness of the memory tree ($M_l$). In Fig.~\ref{figure:lgc} (b-d), we evaluate how these factors influence online video understanding tasks, using the \texttt{Base} model with $t$=0.35.
% with a threshold of $t$=0.35 for consistency across comparisons.
\begin{itemize}[leftmargin=*,itemsep=0pt]\vspace{-2mm}
    \item As shown in Fig.~\ref{figure:lgc} (b), increasing $L$ form 15 to 30 leads to better performance (61.2\%$\rightarrow$64.0\%). However, further increasing $L$ to 40 results in a slight decline (64.0\%$\rightarrow$63.1\%), and substantially increases latency (0.84s$\rightarrow$1.26s), due to the computational demands of the clustering algorithm.
    \item   {Increasing $g$ from 2 to 12, which represents the less compression of visual information and increases input sequence length $C \times L$, enhances performance (62.0\%$\rightarrow$63.9\%)} as greater diversity in the knowledge at each node of the long memory tree $M_l$ is achieved. However, it intensifies the load on the retrieval, leading to an increase in RDP (0.76s$\rightarrow$1.02s), illustrated in Fig.~\ref{figure:lgc} (c).
    \item The clustering goal ($C$) primarily influences the number of tokens ($v_{i}$) stored in the $M_l$. Fig.~\ref{figure:lgc} (d) shows that increasing the dimension of $v_{i}$ (3$\rightarrow$10) enhances model performance (59.4\%$\rightarrow$64.0\%) by enriching the stored knowledge, which also exacerbates VRAM limitations (20$\rightarrow$56 \text{GB}).
    % \vspace{-2mm}
\end{itemize}
% \textbf{First}, as shown in Fig.~\ref{figure:lgc} (b),increasing $L$ form 15 to 30 leads to better performance (61.2\%$\rightarrow$64.0\%). However, this improvement declines when $L$ reaches 40 (64.0\%$\rightarrow$63.1\%), and also causes huge latency (0.84s$\rightarrow$1.26s).
% \textbf{Second}, increasing $g$ from 2 to 12 improves the memory tree structure, leading to better model performance (62.0\%$\rightarrow$63.9\%) due to greater diversity in the knowledge associated with each node. However, this also places more strain on the retrieval algorithm, causing RDP to rise (0.76s$\rightarrow$1.02s), as shown in Fig.~\ref{figure:lgc} (c).
% \textbf{Third}, the clustering goal primarily influences the number of tokens ($v_{i}$) stored in the $M_l$. As shown in Fig.~\ref{figure:lgc} (d), increasing $v_{i}$(3$\rightarrow$10) enhances model performance(59.4\%$\rightarrow$64.0\%) by providing more knowledge, which also leads to increased VRAM shortages (20$\rightarrow$56)GB.

\section{Related Work}
\textbf{Multi-modal Language Models (MLMs).} 
Recent developments of large language models~\cite{brown2020language, zhang2022opt, chung2024scaling, anil2023palm, dubey2024llama, openai2023gpt} and multi-modal alignment techniques have significantly advanced MLMs' capability. 
% Initially, these models focused more on understanding image information. For instance, the BLIP~\cite{li2023blip, panagopoulou2023x} series uses the Q-Former structure to efficiently align image space with text space, while the LLaVA~\cite{liu2024visual, liu2024improved} series employs a simple mapping layer to convert image information and uses visual instruction tuning to enhance the language model’s understanding of images. 
The LLaVA series~\cite{liu2024visual, liu2024improved} utilizes straightforward mapping layers and visual instruction tuning to broaden image understanding tasks to video data. Challenges in video processing primarily involve efficiently compressing video within limited contextual windows. Innovations like ChatUniVi's~\cite{jin2024chat} use of a K-NN clustering algorithm dynamically compress visual tokens, while LLaMA-VID~\cite{li2023llama} reduces single images to two tokens via cross-attention, and MovieChat~\cite{song2024moviechat} leverages long and short-term memory frameworks for extensive data handling. Despite these advances, the transition to effective real-time streaming video understanding in practical applications remains insufficiently addressed. Our research introduces a robust solution designed to meet the real-time demands of online video understanding, aiming to fill this critical gap.
% As advancements in image comprehension have evolved, the scope of these models has broadened to include video information. The primary challenge in processing video is effectively compressing video data within a constrained contextual window. ChatUniVi~\cite{jin2024chat} addresses this by using a K-NN clustering algorithm to dynamically compress visual tokens across temporal and spatial dimensions. LLaMA-VID~\cite{li2023llama} streamlines the representation of single images to just two tokens through cross-attention techniques, while MovieChat~\cite{song2024moviechat} implements a long and short-term memory framework to manage the constraints of extensive video data. Despite these innovations, most current methods remain focused on offline video understanding. The challenge of adapting multi-modal language models to handle real-time streaming video tasks and to store historical information effectively in scenarios that mirror real-world applications is still largely unmet. Our research aims to bridge this gap by introducing a robust solution that caters to the real-time needs of online video understanding.

\textbf{Streaming Video Understanding.} 
Streaming video understanding demands real-time responses from models to user queries, even as video durations potentially extend indefinitely. This is particularly challenging for traditional benchmarks like action recognition~\cite{carreira2017quo}, multi-round video dialogue~\cite{alamri2019audio}, and first-person question answering~\cite{mangalam2024egoschema} which rely on uniform frame sampling. In response to these limitations, there is a growing shift towards online models that process only current and past video frames to formulate responses~\cite{zhang2024flash,chen2024videollm}. 
Despite these advancements, these models often struggle with slow processing speeds and inadequate generalization capabilities, underlining a critical need for further exploration and enhancement in this field.
\textbf{Retrieval-Augmented Generation (RAG).} 
RAG combines information retrieval and text generation to produce more precise and informative responses by incorporating external knowledge into language models~\cite{zhao2024retrieval, ma2023query, yu2022generate, shao2023enhancing, wang2023knowledgpt, dai2022promptagator, sun2022recitation, asai2023self, lin2023ra, ovadia2023fine, chen2023dense}. This technique has become increasingly popular for addressing knowledge retention and real-time information access challenges. MemoryBank~\cite{zhong2024memorybank} 
enhances interaction by storing real-time conversations and leveraging similarity search to retrieve contextually relevant information, enriching the depth and coherence of dialogue. This approach significantly improves a model’s ability to maintain continuity in conversations, particularly in long or multi-turn interactions where maintaining context is crucial.
% enhances interaction by storing real-time conversations and using similarity search to retrieve contextually relevant information, thereby enriching the dialogue with the user. 
Inspired by RAG's efficiency, we introduce a multi-modal memory system that integrates and updates textual and visual data in real time. Using a RAG-inspired retrieval mechanism, this system efficiently accesses the most relevant information from our memory bank, enabling the multi-modal language model to deliver precise, query-specific responses for enhanced video language understanding.

\section{Conclusion}
In this work, we introduce $\StreamBench$, a comprehensive benchmark specifically crafted to assess streaming video understanding, covering a broader range of video lengths and types with six question formats to simulate real-world human-robot interactions. This broader scope enhances our ability to evaluate model performance in complex and dynamic scenarios.
% Compared to previous benchmarks, the proposed $\StreamBench$ covers a broader length distribution for videos and more diverse video types. Among them, we design six different types of questions to more accurately simulate real-world human-robot interactions, offering a richer context for evaluating model performance and aiming to provide broader research possibilities for future studies. 
Alongside, we present $\StreamChat$, a training-free method designed for efficient streaming video understanding, which treats video frames as compressible and storable units and manages them through a hierarchical memory structure.  
With advanced system scheduling, $\StreamChat$ achieves real-time processing speeds and reduced interaction latency, demonstrating robust performance across both online and offline settings in our extensive experiments.

\textbf{Limitations and Future Works.}
Our current retrieval algorithm relies on basic matching techniques, occasionally leading to incorrect responses. Enhancing this with more fine-grained retrieval mechanisms is an essential next step. Additionally, the VRAM constraints of our tree-structured storage could limit scalability as video duration and complexity further grow. Investigating more efficient or adaptive compression techniques will address these limitations. Moreover, to achieve lower latency, we plan to explore closer hardware integration and the potential adoption of fast-serving, multi-modal distributed systems to accommodate larger model parameters and increased user demands.

\bibliography{iclr2025_conference}
\bibliographystyle{unsrt}

\appendix
\newpage
% \section*{Appendix}
% \section{Appendix}
% \begin{center}
%     \large \textbf{SUMMARY OF THE APPENDIX}
% \end{center}
% \vspace{5mm}
\section*{SUMMARY OF THE APPENDIX}
This appendix contains additional details for the ICLR 2025 submission, titled `\textit{Streaming Video Understanding and Multi-round Interaction with Memory-enhanced Knowledge}', which is organized as follows:
% \vspace{-3mm}
\begin{itemize}
\item \S\ref{sec_app:collection_ipeline} presents the dataset collection pipeline.
\item \S\ref{sec_app:visiualize} visualizes more cases of our benchmark.
\item \S\ref{sec_app:retrieval_detail} offers the details about retrieval algorithm.
\item \S\ref{sec_app:metrics_detail} introduces the details and calculation method of metrics.
\item \S\ref{sec_app:failure_case} shows failure cases and analysis.
\item \S\ref{sec_app:selection} {discusses model selection and deployment strategies.}
\item \S\ref{sec_app:expansion} {outlines our plans for benchmark expansion.}
\end{itemize}

\section{Data Pipeline}\label{sec_app:collection_ipeline}
\begin{figure*}[ht]
  \centering
  \setlength{\abovecaptionskip}{0cm}
  \includegraphics[width=0.975\textwidth]{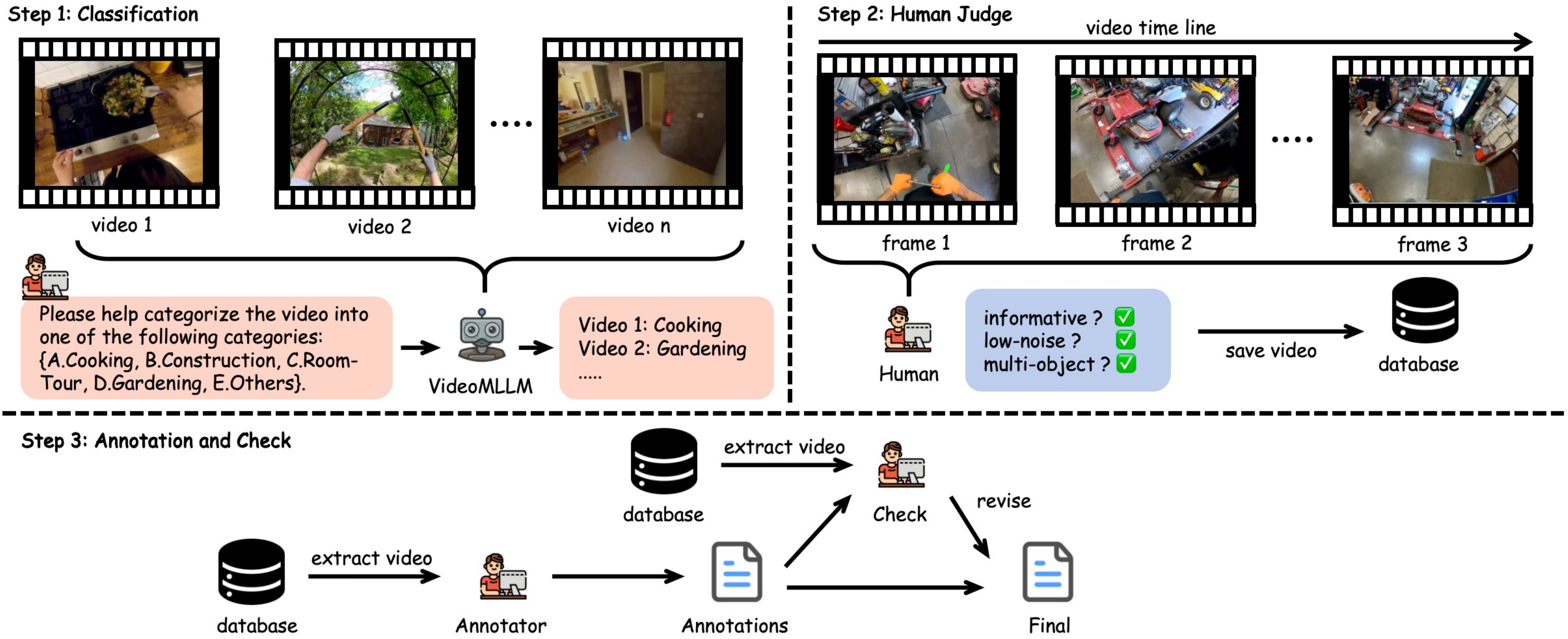}
  % \vspace{-4mm}
  \vspace{2mm}
  \caption{
  The date preparation pipeline utilized in $\StreamBench$.
  }
  \label{figure:video_pipeline}
\end{figure*}

Fig.~\ref{figure:video_pipeline} presents our video collection pipeline.
It consists of 3 parts: (1) Classification; (2) Human judge; and (3) Annotation check.
First, a MLLM~\cite{lin2023video} is utilized to complete the video classification based on our requirements.
{The following prompt is used during the first data filtering step:}

\texttt{`` Based on the observed video information, categorize the video into one of the predefined categories listed in \{\textit{All\_Classes}\}. Respond exclusively in the format of a Python dictionary string with the keys \textit{`pred'} and \textit{`score'}. The `pred' key should contain the uppercase STRING of the chosen category. Refrain from providing any additional text or explanatory output. Your response should strictly follow this example: \{\textit{`pred': `A'}\}.''}
% \end{verbatim}

\texttt{\{\textit{All\_Class}\}} is the options formation.
When dealing with different datasets, we need to change the options.
For example, when dealing with Youtube-8M~\citep{abu2016youtube}, it is \texttt{\{\textit{A: Drama, B: Action, C: Cartoon, D: Romance, E: Sci-fi, F: Others}\}} and \texttt{\{\textit{A: Cooking, B: Construction, C: Room-Tour, D: Gardening, E: Others}\}} for EgoSchema~\citep{mangalam2024egoschema} dataset. We save the output to a JSON file and then find categories from the file as needed.
It is worth noting that we also used the original category information in the YouTube data. The above classification process is used primarily for secondary classification of MovieClips data.
For EgoSchema, we need to classify all original videos as they don't contain category annotations.
% \begin{verbatim}
% ``` Based on the video information you see, please 
% help categorize the video into one of the following categories:
% {All_Classes}. Please generate the response in the form of
% a Python dictionary string with keys 'pred' and 'score', 
% where the value of 'pred' is the uppercase STRING corresponds 
% to the category option. DO NOT PROVIDE ANY OTHER OUTPUT 
% TEXT OR EXPLANATION. Only provide the Python dictionary string. 
% For example, your response should look like this: {'pred': 'A'}.
% '''
%  \end{verbatim}
 
% \{\textit{All\_Class}\} is the options formation.
% When dealing with different datasets, we need to change the options.
% For example, when dealing with Youtube-8M~\citep{abu2016youtube}, it is \{\textit{A: Drama, B: Action, C: Cartoon, D: Romance, E: Sci-fi, F: Others}\} and \{\textit{A: Cooking, B: Construction, C: Room-Tour, D: Gardening, E: Others}\} for EgoSchema~\citep{mangalam2024egoschema} dataset.
% We save the output to a JSON file and then find categories from the file as needed.
% It is worth noting that we also used the original category information in the YouTube data. The above classification process is used primarily for secondary classification of \textit{MovieClips} data.
% For EgoSchema, we need to classify all original videos as they don't contain category annotations.

\section{More Visualizations}\label{sec_app:visiualize}
\begin{figure*}[h]
  \vspace{4mm}
  \centering
  \setlength{\abovecaptionskip}{0cm}
  \includegraphics[width=1.0\textwidth]{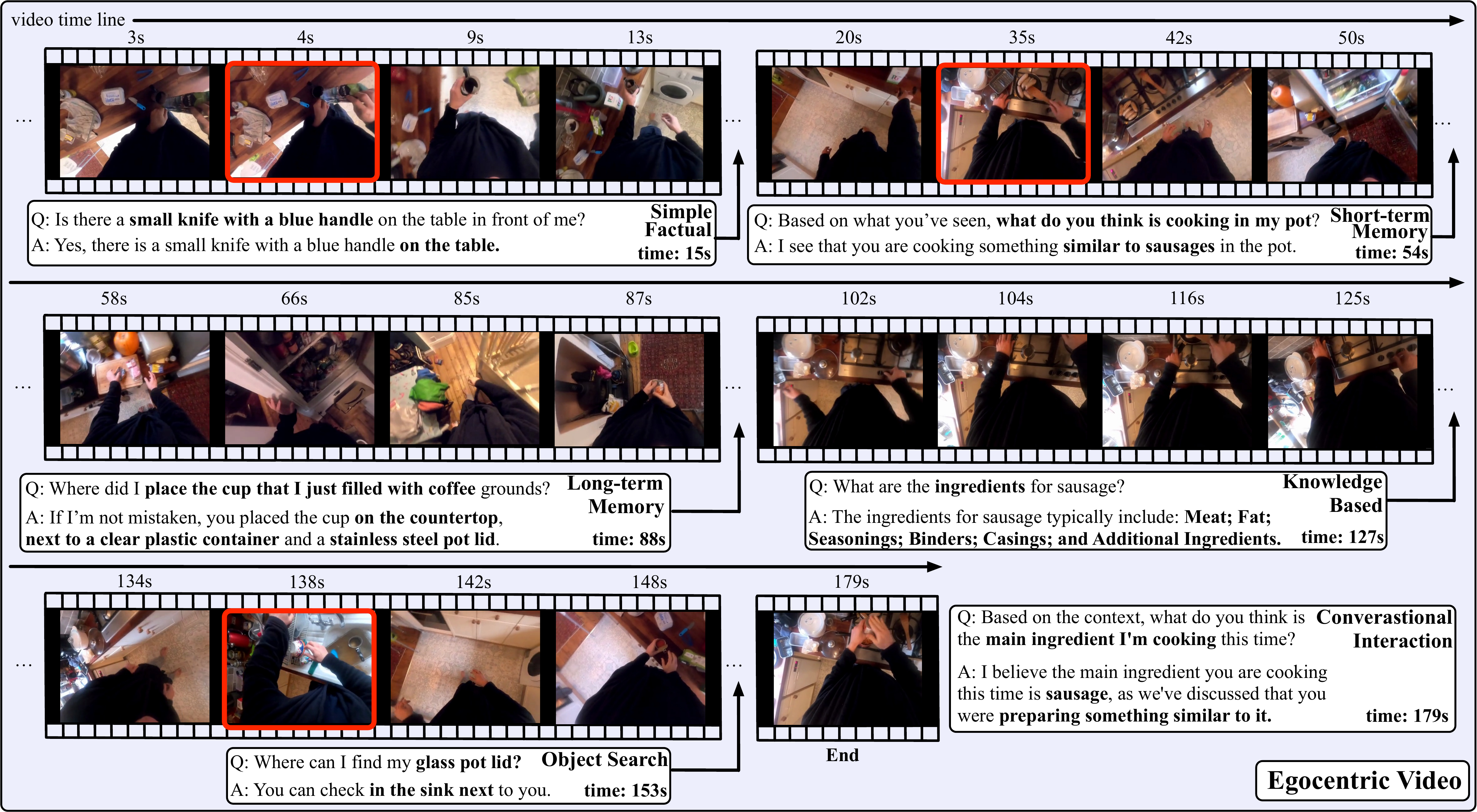}
  % \vspace{-4mm}
  % \vspace{-4mm}
  \vspace{2mm}
  \caption{
  Visualization of egocentric video analysis.
  }
  \vspace{10mm}
  \label{figure:case_ego}
\end{figure*}

\begin{figure*}[ht]
  \centering
  \setlength{\abovecaptionskip}{0cm}
  \includegraphics[width=1.0\textwidth]{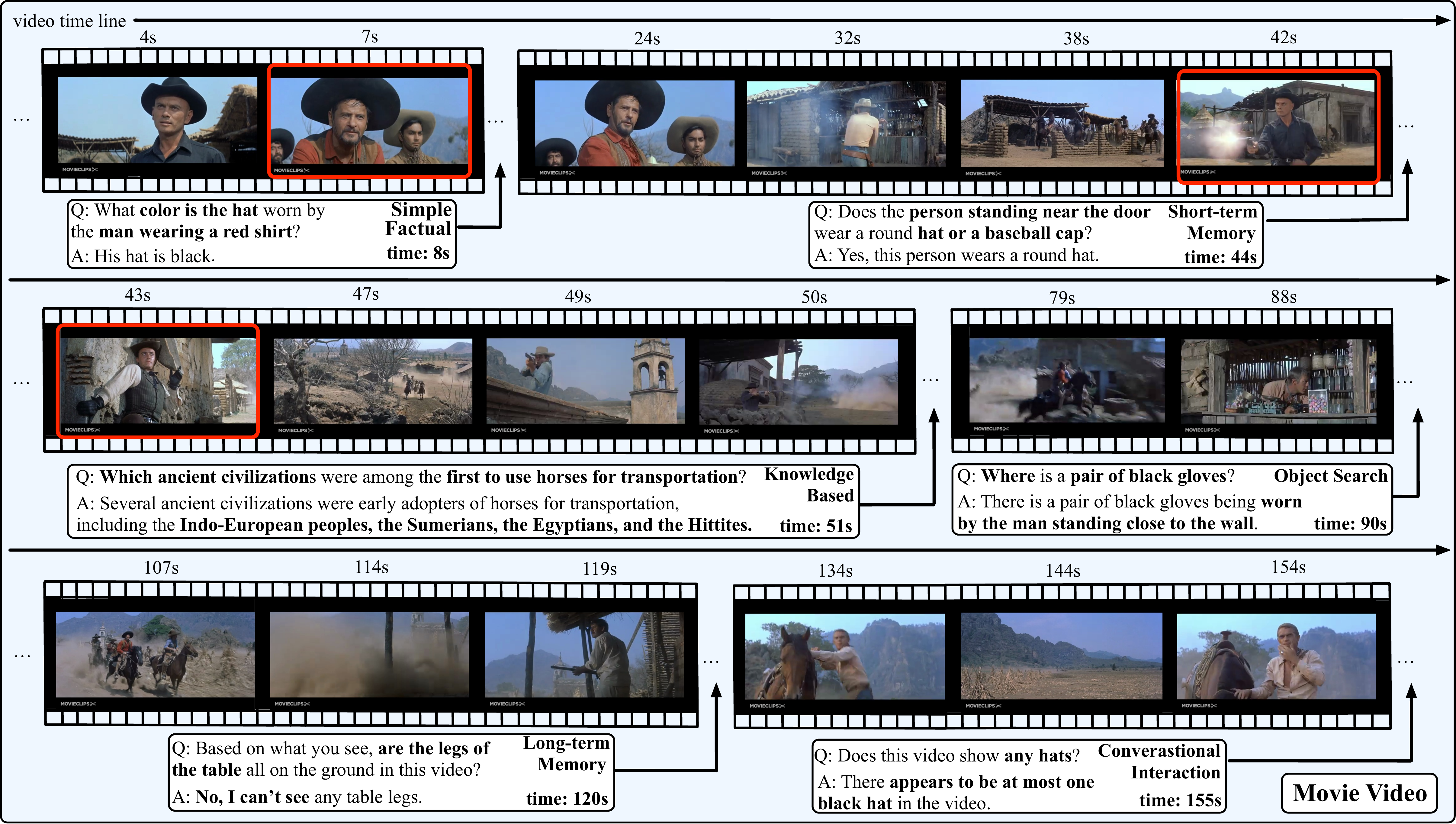}
  % \vspace{-4mm}
  % \vspace{-4mm}
  \vspace{2mm}
  \caption{
 Visualization of movie video analysis.
  }
  \label{figure:case_movie}
\end{figure*}

In Fig.~\ref{figure:case_ego}-\ref{figure:case_movie}, we visualize several $\StreamChat$ cases applied to different types of videos. Specifically, Fig.~\ref{figure:case_ego} illustrates an egocentric video annotation, where the system engages in interactive questioning based on the visual information captured from a first-person perspective. In this figure, various types of annotation questions are showcased, including object identification, memory recall, and knowledge-based interactions. Each example contains six distinct questions, with the Simple Start (SS) question placed first, while the sequence of the remaining five questions varies randomly throughout the video. The questions explore aspects such as short-term memory, long-term memory, object search, and conversational interaction, allowing for a broad range of analysis. To further enhance clarity, specific frames are highlighted along with the locations of key objects referenced in the questions, enabling a better understanding of how the system interacts with the visual context at different moments.

% \begin{figure*}[h]
%   \centering
%   \setlength{\abovecaptionskip}{0cm}
%   \includegraphics[width=1.0\textwidth]{PIC/word_cloud.pdf}
%   % \vspace{-4mm}
%   % \vspace{-4mm}
%   \caption{
%   Word cloud of questions (left) and answers (right) in StreamBench test set.
%   }
%   \label{figure:word_cloud}
% \end{figure*}

\begin{figure*}[h]
  \centering
  \setlength{\abovecaptionskip}{0cm}
  \includegraphics[width=1.0\textwidth]{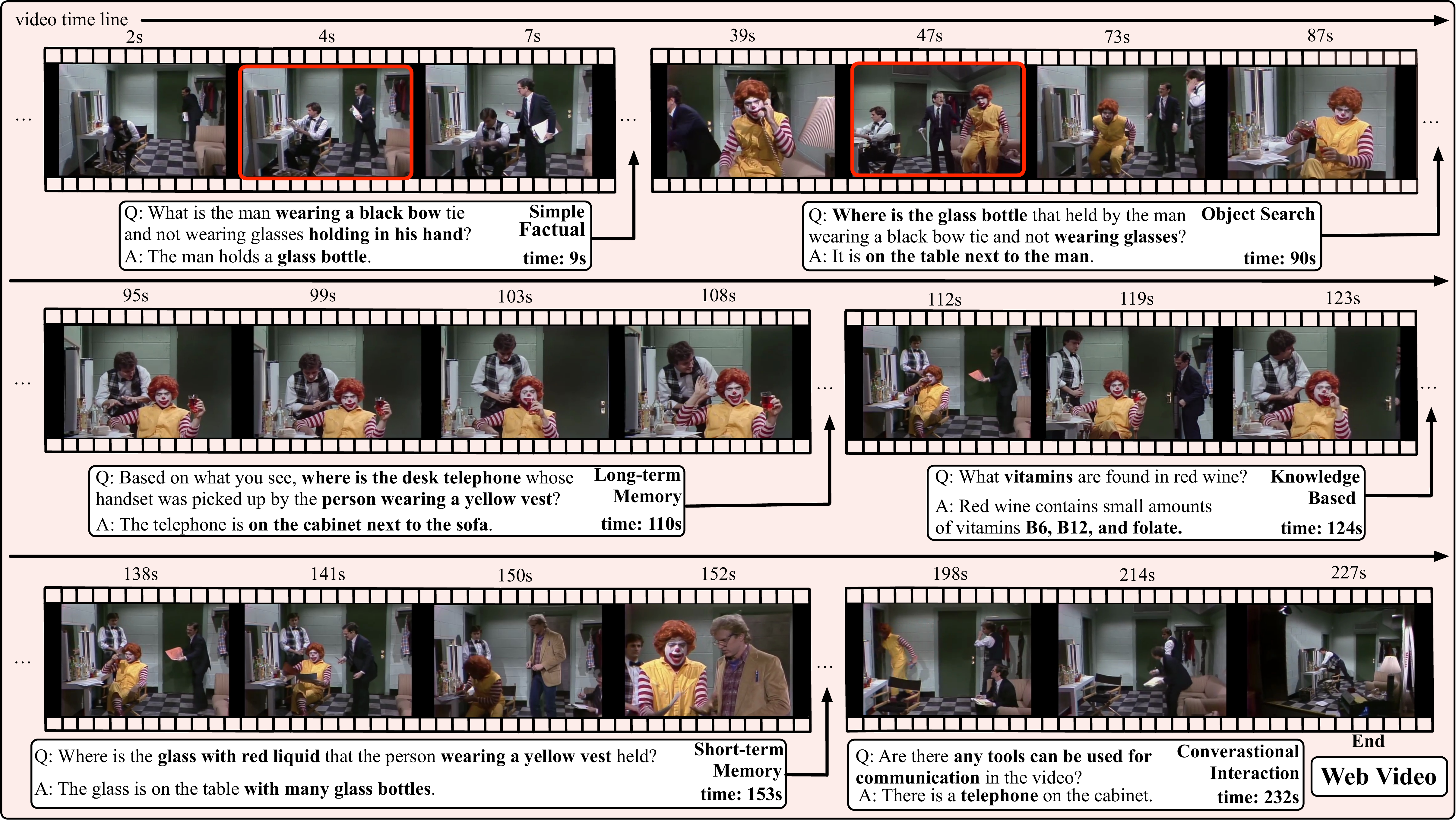}
  % \vspace{-4mm}
  \vspace{2mm}
  \caption{
  Visualization of web video analysis.
  }
  \vspace{10mm}
  \label{figure:case_web}
\end{figure*}

\begin{figure*}[h]
  \centering
  \setlength{\abovecaptionskip}{0cm}
  \includegraphics[width=1.0\textwidth]{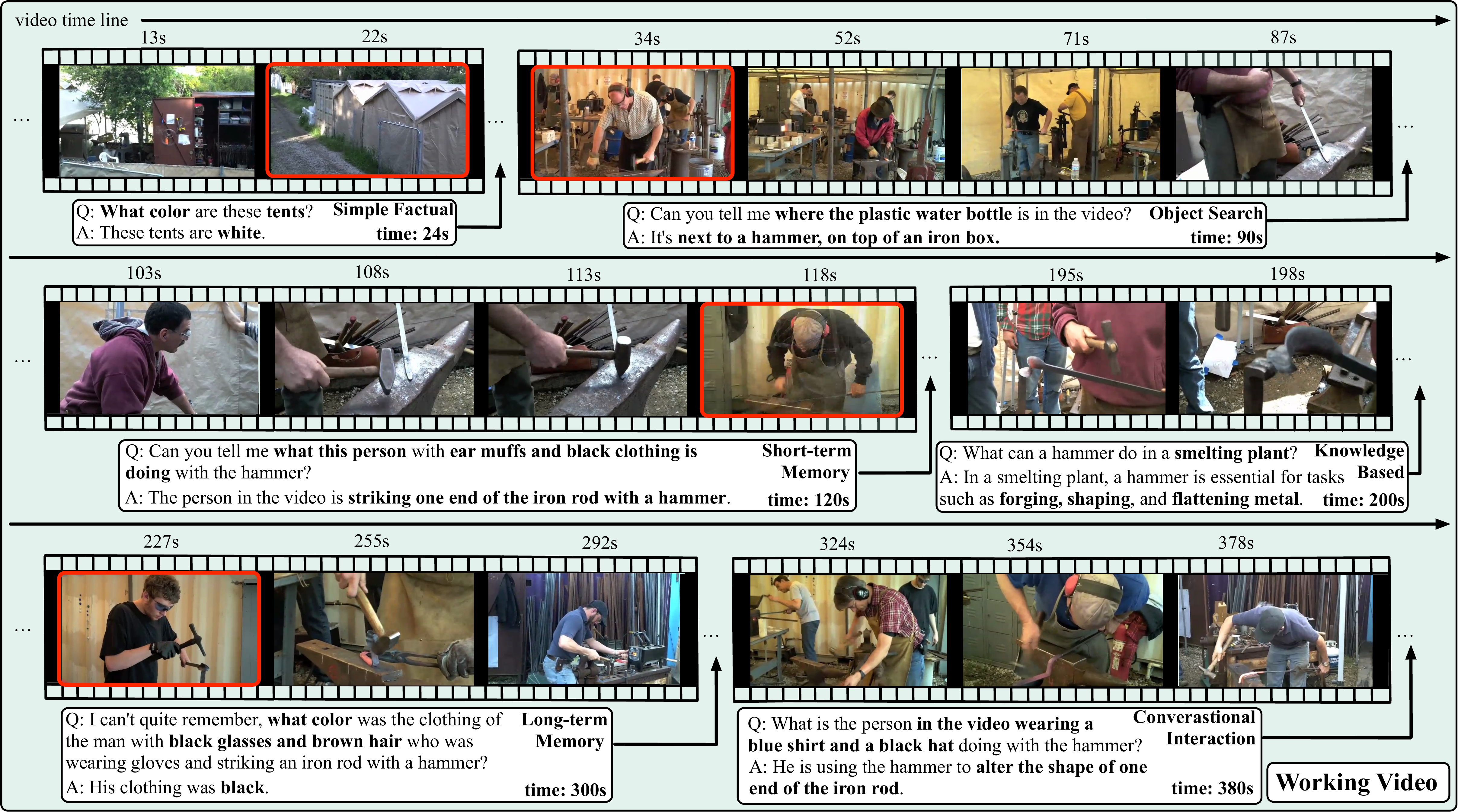}
  % \vspace{-4mm}
  % \vspace{-4mm}
  \vspace{2mm}
  \caption{
  Visualization of working video analysis.
  }
  % \vspace{mm}
  \label{figure:case_web}
\end{figure*}

\section{Retrieval Algorithm}\label{sec_app:retrieval_detail}
Inspired by the retrieval argumentation system~\cite{zhao2024retrieval}, our approach enhances the model's capability to address complex queries by retrieving the most relevant information from long-term memory for contextual support. As outlined in Algorithm 1, we compute the similarity, \text{Sim}, between the user's request, $Q$, and entries $T_n$ in the memory. This process identifies the optimal path for accessing the most pertinent stored knowledge, $T_{best}$ and $C_{best}$. Leveraging our tree-like storage structure, we efficiently focus on the highest-similarity nodes at each layer, minimizing the computational load by avoiding exhaustive sub-node calculations. The selected knowledge, $T_{best}$ and $C_{best}$, is then integrated with $Q$ in a mixed prompt format to serve as the final input for the multi-modal language model, facilitating accurate response generation.
\vspace{2mm}
% Inspired by the retrieval argumentation system~\cite{zhao2024retrieval}, when facing more challenging questions, we need to retrieve the most relevant information from long-term memory as a contextual supplement to help the model generate the correct answer. 
% Algorithm~\ref{algorithm:knowledge} shows the calculation process.
% By calculating the similarity  \text{Sim}  between the user’s request $Q$ and $T_n$, we can find the optimal path in the memory storage and extract the stored knowledge $T_{best}$  and $C_{best}$ from the optimal path. Thanks to the tree-like storage structure, we only need to focus on the nodes with the highest similarity at each layer, avoiding the time consumption caused by calculating all sub-nodes. The retrieved knowledge $T_{best}$ and $C_{best}$ are then recombined with $Q$ and used as the final input in a mixed prompt format for the MLLM.

\begin{algorithm}
\caption{Knowledge Retrieval from Long-Term Memory}
\begin{algorithmic}[1]
\State \textbf{Input:} User request $Q$
\State \textbf{Output:} Best matching knowledge $T$ and $C$
\State Initialize similarity $Sim_{max} \gets -\infty$
\State Initialize best path knowledge $T_{best}, C_{best} \gets \emptyset, \emptyset$
\For{each node $n$ in the tree structure}
    \State Compute similarity $Sim(Q, T_n)$ where $T_n$ is the caption at node $n$
    \If{$Sim(Q, T_n) > Sim_{max}$}
        \State $Sim_{max} \gets Sim(Q, T_n)$
        \State $T_{best}, C_{best} \gets T_n, C_n$ \Comment{Update best match knowledge}
    \EndIf
    \If{node $n$ has children}
        \State Continue to next level
    \EndIf
\EndFor
\State Reconstruct input for MLLM using $T_{best}$, $C_{best}$, and $Q$
\State \Return $T_{best}$, $C_{best}$
\end{algorithmic}
\end{algorithm}

\section{Details of Metrics}\label{sec_app:metrics_detail}
We design the following metrics to measure the model’s ability to stream video understanding:

\textbf{(1) Score and Accuracy}:
To assess the semantic correctness of a single-turn dialogue, using language models is a mainstream approach~\cite{xu2017video, xu2016msr, yu2019activitynet, fu2024video, zhou2024mlvu}. We also use this as a key metric in our benchmark. In our test benchmark, we use the open-source language model LLaMA-3 8B ~\cite{dubey2024llama} Instruct version as our scoring model $f$. 
{Here is the prompt that we used during scoring:}

\begin{table}[h!]\centering
\vspace{2mm}
\begin{minipage}{0.99\columnwidth}\vspace{0mm}    
\centering
\begin{tcolorbox} 
    \centering
    \small
     \hspace{-6mm}
    \begin{tabular}{p{0.99\columnwidth}}
% \begin{minipage}{0.99\columnwidth}\vspace{0mm}
\VarSty{Prompt} = [
    \{\var{"role": "system", "content":} "You are an intelligent chatbot designed for evaluating the correctness of generative outputs for question-answer pairs.\\
    Your task is to compare the predicted answer with the correct answer and determine if they match meaningfully. 
    Here's how you can accomplish the task: \\
    INSTRUCTIONS: \\
    - Focus on the meaningful match between the predicted answer and the correct answer. \\
    - Consider synonyms or paraphrases as valid matches. \\
    - Evaluate the correctness of the prediction compared to the answer.
    "\} \\
    \{\var{"role": "system", "content":} "Please evaluate the following video-based question-answer pair:
    Question: {{question}}; Correct Answer: {{answer}}; Predicted Answer: {{prediction}} \\
    Provide your evaluation only as a yes/no and score where the score is an integer value between 0 and 5, with 5 indicating the highest meaningful match. \\
    Please generate the response in the form of a Python dictionary string with keys 'llama pred' and 'score', where the value of 'llama pred' is a string of 'yes' or 'no' and the value of 'score' is in INTEGER, not STRING.
    DO NOT PROVIDE ANY OTHER OUTPUT TEXT OR EXPLANATION. Only provide the Python dictionary string. 
    For example, your response should look like this: \{'llama pred': YOUR JUDGE, 'score': YOUR SCORE.\}"
    \}
    ]
% \end{minipage}
\end{tabular}
\end{tcolorbox}
\vspace{-2mm}
\caption{
{Prompt given to the LLaMA-3 model for evaluation.}
}
\label{tab:judgment prompt}
\end{minipage}
\vspace{2mm}
\end{table}

{We organize the question $Q$, reference answer $R$, and model’s response $M$ into the Tab.~\ref{tab:judgment prompt} formation and send to scoring model,} which then provides a score in the range of 0-5 and evaluates whether the model’s response is semantically correct:
\begin{equation}
    S_i= f(Q, R, M), Acc = \frac{1}{N} \sum_{i=1}^{N} \mathbb{I}(S_i \geq T)
\end{equation}
A higher score $S_i$ and $Acc$ indicates that the answer is closer to the reference answer.

\textbf{(2) Coherence}:
Given that a single video may involve multiple rounds of dialogue, we need to evaluate the model’s ability to provide a coherent experience across different rounds. We introduce the coherence metric, which calculates the absolute value of the difference between the semantic scores $S_i$ of different dialogues within a single scenario. The average of all these differences is used as the coherence metric. The calculation formula is as follows:
\begin{equation}
    C = \frac{1}{N-1} \sum_{i=1}^{N-1} \left| S_i - S_{i+1} \right|,
\end{equation}

where \( C \) is the coherence score, \( N \) is the total number of dialogue turns in the scenario, \( S_i \) represents the semantic score of the \(i\)-th dialogue turn and \( \left| . \right| \) is the absolute difference between the semantic scores of consecutive dialogue turns.
It is evident that a smaller \( C \) indicates that the model provides a better coherence experience for the user.

\textbf{(3) Request Processing Delay}:
For online scenarios, system latency consists of two parts: 1. Request processing delay; 2. Generation delay. The generation delay is mainly influenced by factors such as context length, language model parameters, and deployment methods, and can be adjusted through various methods. In this benchmark, we primarily assess (1) request processing delay, which is calculated as the time from when the user completes the request input to when the model starts generating the response. The calculation formula is as follows:
\begin{equation}
    \text{RPD} = T_{\text{start}} - T_{\text{input}}
\end{equation}

\section{Failure Case and Analyze}\label{sec_app:failure_case}
We present some failure cases that occurred during testing and explain why they occurred.
Most of these cases come from object search, long-term memory, short-term memory and conversational interaction tasks.
The problems that occurred are mainly grouped into four types:
\begin{itemize}[leftmargin=*,itemsep=0pt]
\item \textbf{Temporal Fine-grained:}
% In the object search task, our method still has difficulty finding key information if the objects or events we ask about appear too briefly or discretely. For example, as shown in case 1, the user's question is about the object candle, but because the frequency of candles is too rare and the object itself is small, the model cannot answer correctly.
In the object search task, our method still struggles to identify key information when the queried objects or events appear too briefly or sporadically. For instance, as demonstrated in case (1), the user’s question pertains to a candle. However, due to the infrequent appearance of the candle and its small size, the model fails to provide an accurate response.
\item \textbf{{Spatial Fine-grained:}} 
% Whether in long-term or short-term memory, if the user's target is too small or cannot stand out from the background, our method still has limitations even for objects that appear multiple times in the video. For example, in case 2 and case 3, the target cup and bowl are too small relative to the foreground, causing the model to be unable to find the target.
{Whether in object search or short-term memory task, our method faces limitations when the user’s target is too small or blends into the background, even if the object appears multiple times in the video. For example, in short-term memory task case (2) and object search task case (3), the target objects (porcelain bowl and red cup) are too small relative to the foreground, making it difficult for the model to accurately detect and locate them.
We will continue to improve our method to enhance the perception of small objects.
}
\item \textbf{Target Movement:} 
% In some scenarios, the user's questions will involve task action inference. 
% During the reasoning process, we found that in some cases, even if the model determined the target, its judgment of its action and the relationship with the surrounding objects were still wrong. The model could not determine the action association between the "person" and the "box", resulting in an inability to answer correctly.
During the reasoning process, we observed that in some cases, even when the model correctly identified the target, its interpretation of the target’s actions and relationships with surrounding objects was still inaccurate. For instance, in the long-term memory task case (4) the model failed to recognize the action and position association between the “person” and the “box,” leading to an incorrect response.
\item \textbf{Context Induction:} 
% In the conversational interaction task, the model will be affected by the accuracy of the answers to related questions. In case 5, the model retrieved the previous dialogue history, but when the historical information is wrong, it is difficult for the model to answer correctly.
% This is because the context information will have an irreversible influence on the model's judgment, resulting in the model being unable to answer correctly.
In the conversational interaction task, the model’s performance is influenced by the accuracy of its responses to previous related questions. For instance, in case 5, the model retrieved information from the dialogue history, but when that historical information was incorrect, it became challenging for the model to provide the correct answer.
\item \textbf{{Information Loss:}} {
According to the experimental results in Tab.~\ref{tab:types_score}, although our method shows balanced performance in various tasks of StreamChat, our hierarchical memory storage still has the potential risk of losing information. Since our existing method is too dependent on the accuracy of the retrieval algorithm, we will continue to update our method to minimize information loss.
}
\end{itemize}
% \textbf{Fine-grained Retrieval}:
% In order to further improve the performance of the model in online scenarios, we believe that more fine-grained information retrieval is the key. Our retrieval algorithm is still a simple matching method at present, and there is still the possibility of matching errors, which will directly cause the model to go to the wrong memory branch and fail to generate the correct answer.

% \textbf{VRAM budget}:
% Although our method can achieve compression and storage of video information, at present, the tree-structured storage method is still limited by VRAM as discussed in Sec.~\ref{Ablation}. As the video time increases or the memory parameters change, our method is at risk of exceeding the existing limit.
% We may need to find more efficient or learnable compression methods to solve this problem.

% \textbf{Lower Latency}:
% System scheduling is an engineering solution to reduce latency. We think achieving lower latency may need to work closely with the hardware system to handle larger model parameters and achieve efficient inference for more user requests. For example, a more complete multi-modal distributed system or a fast-serving framework similar to SGLang~\citep{zheng2023efficiently} and vLLM~\citep{kwon2023efficient}.
% \vspace{-4mm}
\begin{figure*}[h]
  \centering
  % \vspace{-4mm}
  \setlength{\abovecaptionskip}{0cm}
  \includegraphics[width=0.95\textwidth]{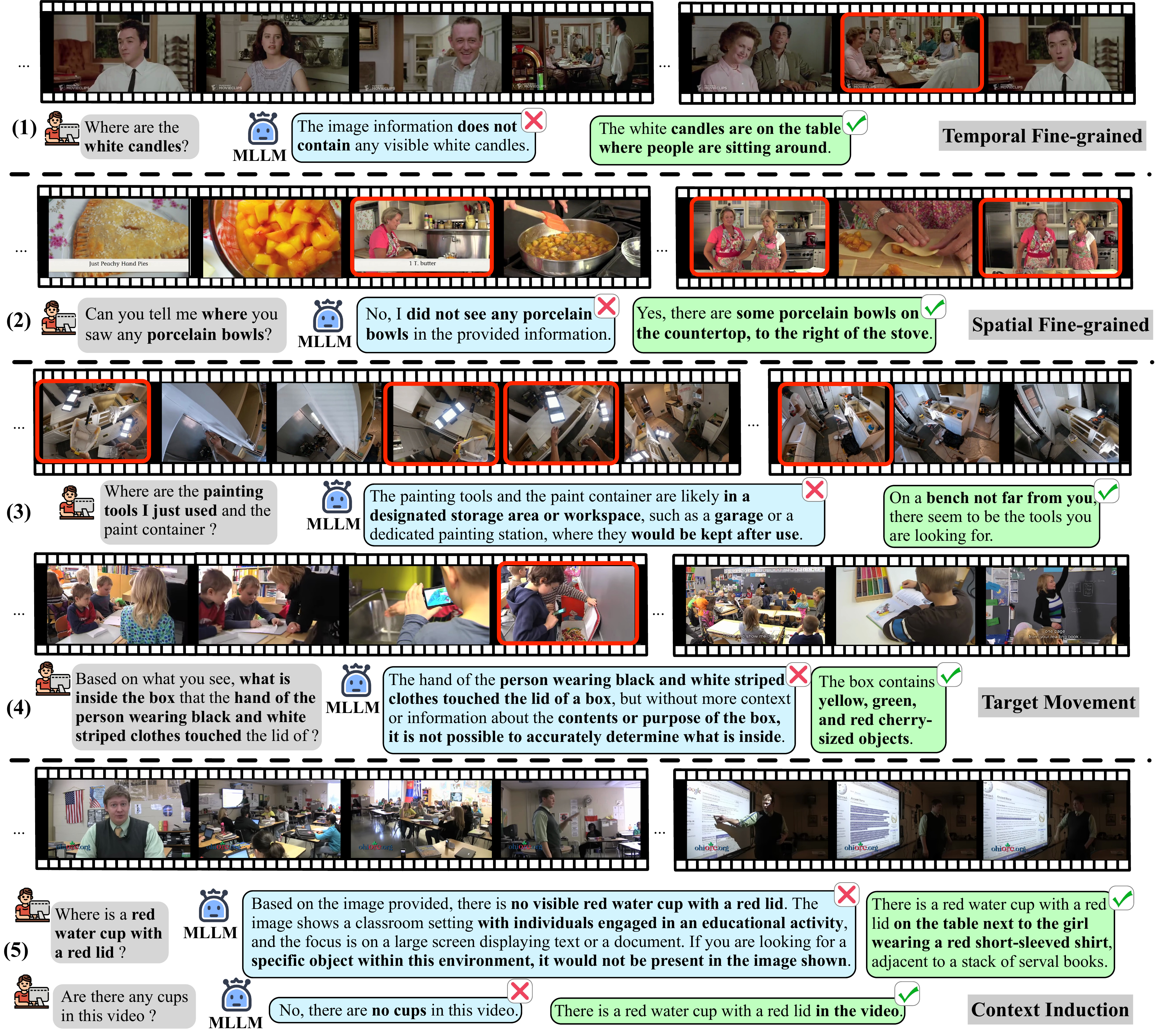}
  % \vspace{-4mm}
  % \vspace{-4mm}
  \vspace{1mm}
  \caption{
  Visualization of failure cases.
  }
  \vspace{-2mm}
  \label{figure:faliure_case}
\end{figure*}

\vspace{-4mm}
\section{Model selection and deployment}\label{sec_app:selection}
Our research indicates that a suitable model should possess the following key attributes:
\begin{itemize}[leftmargin=*,itemsep=0pt]
\item \textbf{Long-Form Video Understanding:} Effective processing of long videos is crucial. While we utilize K-Means for feature compression, the information retrieved by our memory mechanism remains relatively long, requiring a model capable of handling extended sequences.
\item \textbf{Robustness to Prompt Variations:} For accurate and coherent multi-turn conversations, the model must be robust to changes in prompt wording.  This robustness is essential to prevent inconsistencies or hallucinations in the model's output when prompts are adjusted to incorporate information from the memory mechanism.
\end{itemize}
By integrating LongVA with our proposed system, we successfully extend its capabilities to encompass streaming video processing and multi-turn conversations while preserving these critical characteristics.
As we introduced in \S\ref{exp}, we utilize 2 GPUS to complete the deployment of our method. The main reason is that during system scheduling, we need to utilize tensor parallelism to distribute the computational load for efficient execution. Specifically, the \textit{(i) selective frame stacking thread} and \textit{(iii) context summarization thread} are running on GUP1 while \textit{(ii) memory formation thread} is running on GPU2.Therefore, the compressed video tensors need to be transmitted between different GPUs to ensure the stable operation of the system.

% \vspace{-8mm}
\section{Expansion plan of StreamBench}\label{sec_app:expansion}
\begin{itemize}[leftmargin=*,itemsep=0pt]
\item \textbf{Video scale:} We are trying to expand the number of videos contained in the StreamBench to reach a higher standard. We are working on expanding the number of videos to \textbf{thousands} while maintaining the diversity of video length and types.
\item \textbf{Annotation scale:} We are continuing to promote the development of high-quality annotations. Based on your suggestions, we will use manual annotation methods to expand the annotation of existing benchmarks to the order of \textbf{ten thousand levels} and also use manual inspection to filter out toxic labels and erroneous information.
\item \textbf{Diverse tasks:} Given that the current benchmark only has a single task type, we are continuing to expand the types of tasks included in the benchmark, including but not limited to \textbf{multiple-choice questions, video captioning, and video grounding and etc.}
\end{itemize}

\end{document}